\pdfoutput=1

\documentclass[11pt]{article}

\usepackage{latex/acl}

\usepackage{times}
\usepackage{latexsym}
\usepackage{tcolorbox}
\usepackage[T1]{fontenc}
\usepackage{subcaption}
\usepackage[utf8]{inputenc}

\usepackage{microtype}

\usepackage{inconsolata}
\usepackage{amssymb}
\usepackage{amsmath}
\usepackage{booktabs}
\usepackage{hyperref}
\usepackage{multirow}
\usepackage{graphicx}
\usepackage{makecell}

\newcommand{\modelname}[0]{CItruS}

\title{\modelname{}~\includegraphics[width=1em]{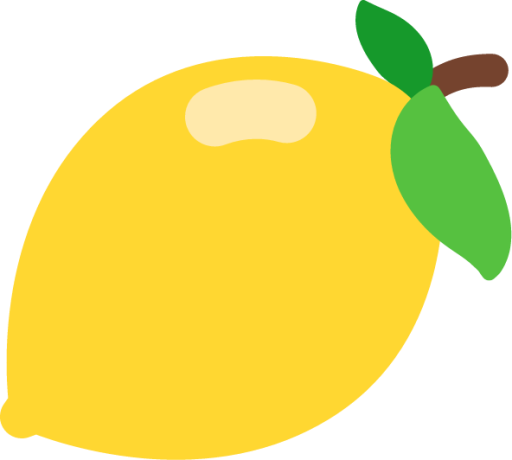}: Chunked Instruction-aware State Eviction\\ for Long Sequence Modeling}

\author{Yu Bai$^{1,2}$$^*$, Xiyuan Zou$^{3,4}$\thanks{~~Equal contribution. Work done during the internship at Mila – Quebec Artificial Intelligence Institute.}, Heyan Huang$^{1,2}$~\thanks{~~Corresponding author.},\\ \textbf{Sanxing Chen$^6$, Marc-Antoine Rondeau$^3$, Yang Gao$^1$, and Jackie Chi Kit Cheung$^{3,4,5}$}  \\
    $^1$Beijing Institute of Technology, Beijing, China\\
    $^2$Southeast Academy of Information Technology, Fujian, China\\
    $^3$Mila – Quebec Artificial Intelligence Institute\\
    $^4$McGill University ~~ 
    $^5$Canada CIFAR AI Chair ~~
    $^6$Duke University\\
  \texttt{yubai@bit.edu.cn, xiyuan.zou@mail.mcgill.ca} \\
    }

\begin{document}
    \maketitle
\begin{abstract}

Long sequence modeling has gained broad interest as large language models (LLMs) continue to advance. 
Recent research has identified that a large portion of hidden states within the key-value caches of Transformer models can be discarded (also termed \emph{evicted}) without affecting the perplexity performance in generating long sequences.
However, we show that these methods, despite preserving perplexity performance, often drop information that is important for solving downstream tasks, a problem which we call \emph{information neglect}.
To address this issue, we introduce \textbf{C}hunked \textbf{I}ns\textbf{tru}ction-aware \textbf{S}tate Eviction (\textbf{\modelname{}}), a novel modeling technique that integrates the attention preferences useful for a downstream task into the eviction process of hidden states. 
In addition, we design a method for chunked sequence processing to further improve efficiency.
Our training-free method exhibits superior performance on long sequence comprehension and retrieval tasks over several strong baselines under the same memory budget, while preserving language modeling perplexity. 
The code and data have been released at  \url{https://github.com/ybai-nlp/CItruS}.

\end{abstract}

\section{Introduction}
\label{sec:intro}

Recent advances in large language models (LLMs) have raised interest in long sequence modeling~\citep{qin-etal-2023-nlp, xiao2023efficient}. Several studies have found that information relevant to the next token prediction task often accumulates in the hidden representations of just a few tokens, and the attention distributions tend to focus sparsely on these tokens~\citep{liu2024scissorhands, bai2024analyzing, wang2023label}. This observation has resulted in methods that model longer sequences by evicting unnecessary key-value caches during the language modeling process~\citep{zhang2024h2o, oren2024transformers}, mostly based on the attention weights each token receives from the following context.

\begin{figure}[t]
\centering
\includegraphics[width=\columnwidth]{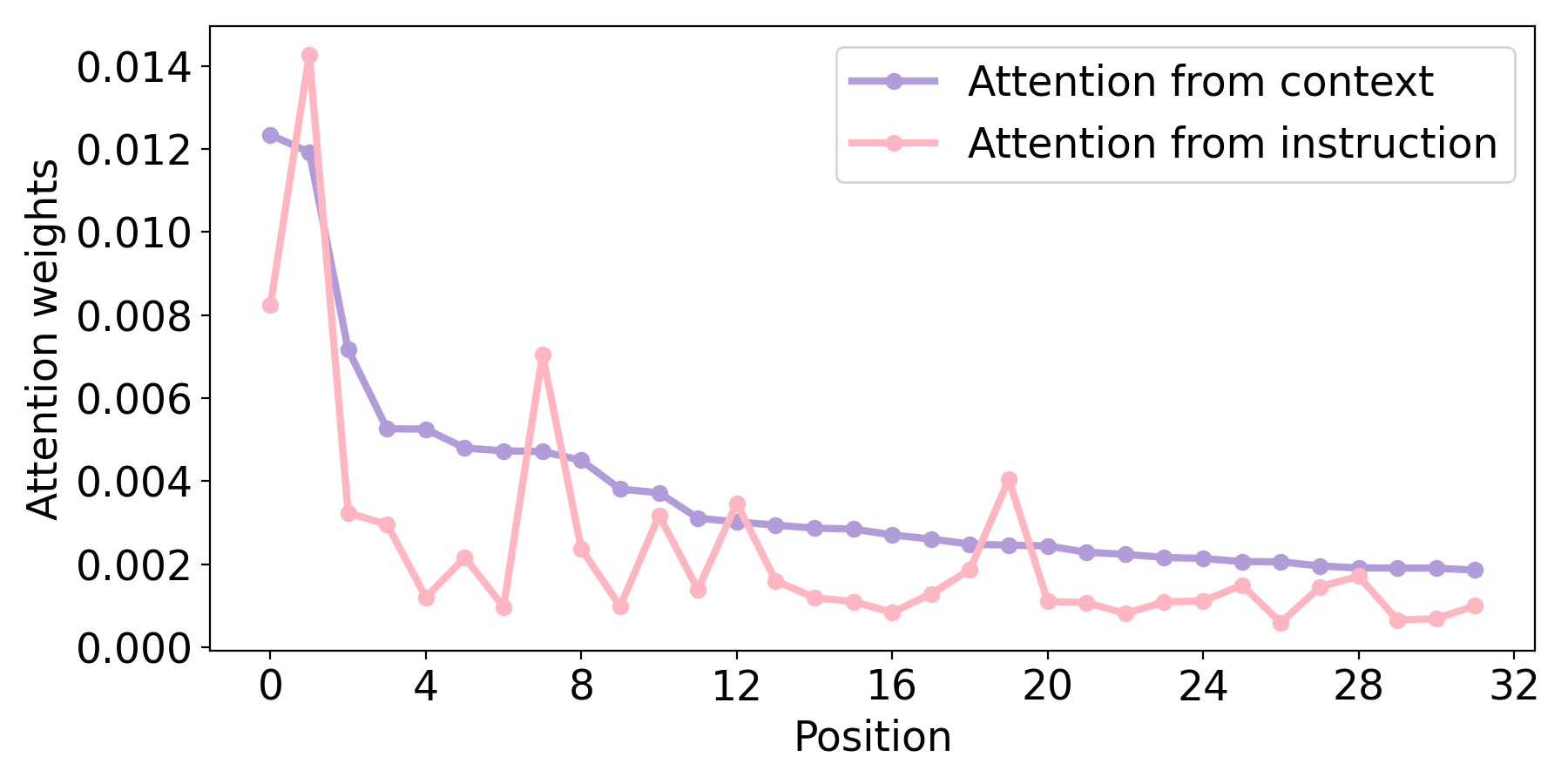} 
\caption{One sample from attention distributions in the 16th layer of the Mistral 7B Instruct model applied to the Qasper dataset. The attention distributions are calculated from a document context and an instruction text to the key-value cache. The x-axis represents different positions within the key-value cache, while the y-axis represents the attention weights. The positions are reordered by descending attention weights from the context, and positions with low attention weights are omitted for clarity.}
\label{fig:intro_attention}
\end{figure}

However, these methods achieve limited performance on downstream tasks that require specific information from long documents (e.g., question answering), suggesting that they struggle to retain the detailed information necessary for such tasks. 
We refer to this condition as the \emph{information neglect} problem.
This issue arises because the cache acquired through state eviction is based only on the local document context. There is no explicit signal for the model to ensure that it is useful for solving downstream tasks.
Consider Figure~\ref{fig:intro_attention}, which shows two attention distributions---one from a document context and one from an instruction prompt---when applying the Mistral 7B Instruct model to a sample from the Qasper dataset. Note that the two differ substantially in their weighting of positions, suggesting that the document context-derived attention weights may not capture well the task specified by the instructions.\footnote{More details are provided in Section~\ref{sec:information_neglect}.} 


In this paper, we propose to address this information neglect issue by incorporating the instruction text into the state eviction process. Our method, \textbf{C}hunked \textbf{I}ns\textbf{tru}ction-aware \textbf{S}tate Eviction~(\textbf{\modelname{}}), 
decomposes long sequence processing into two different subprocesses: language modeling and task solving.
For the language modeling process, we propose chunked state eviction, splitting the long sequence input into large chunks while maintaining a cache that only stores the most important key-value states, which we show allows the model to efficiently and effectively encode long documents. As for the task-solving process, we propose an instruction-aware cache, either independent of or shared with the language modeling cache, which maintains the specific detailed information required to generate responses in downstream settings. The instruction-aware cache is then used to generate the final response for solving the task.
Our approach can be seen as analogous to ideas from cognitive science that language and thought can be disentangled in human language processing~\citep{fedorenko2016language},

We evaluate \modelname{} on three tasks: long document reading comprehension, knowledge retrieval, and language modeling. Our approach improves downstream task performance over several strong baselines by large margins and enables the retrieval of desired information hidden within a long document of up to one million tokens. Furthermore, the model maintains high language modeling performance with a low perplexity. Notably, \modelname{} is applicable to all the transformer-based decoder-only model without any further training, improving the model's ability to conduct downstream tasks for input sequences with arbitrary lengths.

Overall, our contributions are summarized as follows: 
    1)  We define and demonstrate the information neglect problem in state-eviction methods.
    2) We propose \modelname{}, a state eviction method designed for long sequence downstream tasks, which incorporates an instruction-aware cache for task-solving and a chunked state eviction process for efficient language modeling.
    3) Experiments on long document reading comprehension, knowledge retrieval, and language modeling show that \modelname{} improves performance on downstream tasks involving long sequence by large margins while maintaining low language modeling perplexity.

\section{Related Work}
\subsection{Long Sequence Processing}
Long sequence processing has long been a key research area in natural language processing~\citep{tiezzi2024resurgence}. Various approaches have been explored to address this challenge, including Longformer and State Space Models~\citep{beltagy2020longformer, gu2021efficiently, gu2023mamba}. Additionally, memory-augmented models use external memory to handle long sequences~\citep{56193,wu2022memorizing, bertsch2024unlimiformer, lu2024longheads}, while recurrent-based transformers have been designed for long-sequence tasks~\citep{dai2019transformerxl, li-etal-2023-recurrent, peng2023rwkv}. 
More related work about long sequences is further provided in Appendix~\ref{app:related_work}.




Except for \textsc{LongHeads}, a memory-augmented method which requires storing all the past key-value states, all the above methods require further training of the model to handle long sequence processing. Our approach is an inference-time method and eliminates the need for further training, working directly with any open-source transformer-based language model and requiring significantly fewer resources than the methods mentioned.

Our work is also similar to retrieval-augmented generation (RAG) methods~\citep{gao2023retrievalaugmented, zhao2024retrievalaugmented}, which incorporates knowledge from external databases to enhance the generation. 
However, RAG research mainly focuses on the retrieval process in order to better leverage the documents that could support the response generation, whereas \modelname{} is a method that more generally focuses on performing various long sequence tasks.  It could be a good option to be applied to the RAG process. 
In fact, our testing includes long-document question answering and retrieval as primary tasks.
We further discuss the difference between our method and RAG in Appendix~\ref{app:related_work}.

\subsection{State Eviction for Large Language Models}
\citet{liu2024scissorhands} explore the persistence of importance hypothesis for the key-value cache of large language models, which states that the position of the cache that are useful for language modeling tend to remain consistent over time. Based on this, various methods that evict the key-value cache during language modeling have been proposed for improving the efficiency of LLM inference.
\citet{zhang2024h2o} use accumulative attention scores to evict unnecessary key-value cache states.
\citet{oren2024transformers} use the attention of the last token as a metric for evicting hidden states.
\citet{ge2023model} profile all the attention heads and maintain different hidden states for different heads.
\citet{ren2024efficacy} propose determining the eviction scope by evaluating the standard variance of the attention weights received by individual tokens, and they test the efficiency improvement of state eviction methods using small text chunks of size 16, which we scale up to 768 in our work.
\citet{yang2024attendre} bring the preference of future tokens into the state eviction process.
\citet{xiao2023efficient} propose that ``attention sinks'' exist during LLM sequence processing. By keeping the key-value states of the initial tokens, and evicting the key-value states out of a sliding window maintained for recent tokens, their model could maintain the perplexity while processing 4 million tokens.


We propose that these previous methods suffer from the information neglect problem; that is, they fail to preserve specific information related to the instruction text, therefore might lower the performance on down-stream tasks.

\section{The Information Neglect Problem}
\label{sec:information_neglect}
In this section, we demonstrate the information neglect problem of existing state eviction methods.
State eviction methods have two basic elements: a key-value cache $C$ that maintains the most important hidden states for language modeling and a strategy $S$ to evict unnecessary states from the key-value cache, thereby making room for new states. By iteratively evicting the most unnecessary tokens from the cache, the model achieves the capability to model long sequences of arbitrary lengths. $S$ is usually based on the attention weight a cache state receives from tokens later in the sequence.

\begin{figure}[t]
\centering
\includegraphics[width=0.95\columnwidth]{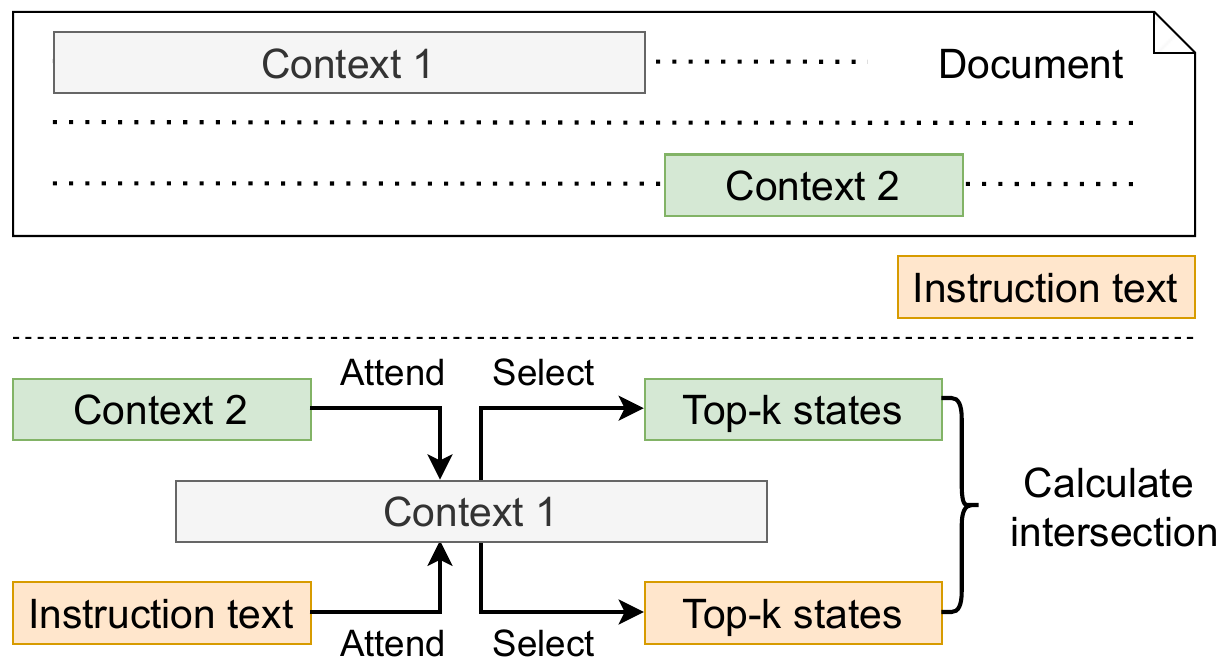} 
\caption{The illustration of our experiments that apply intersection calculation to explore the information neglect problem in state eviction models.}
\label{fig:overlap_method}
\end{figure}

\begin{figure}[t]
\centering
\includegraphics[width=0.95\columnwidth]{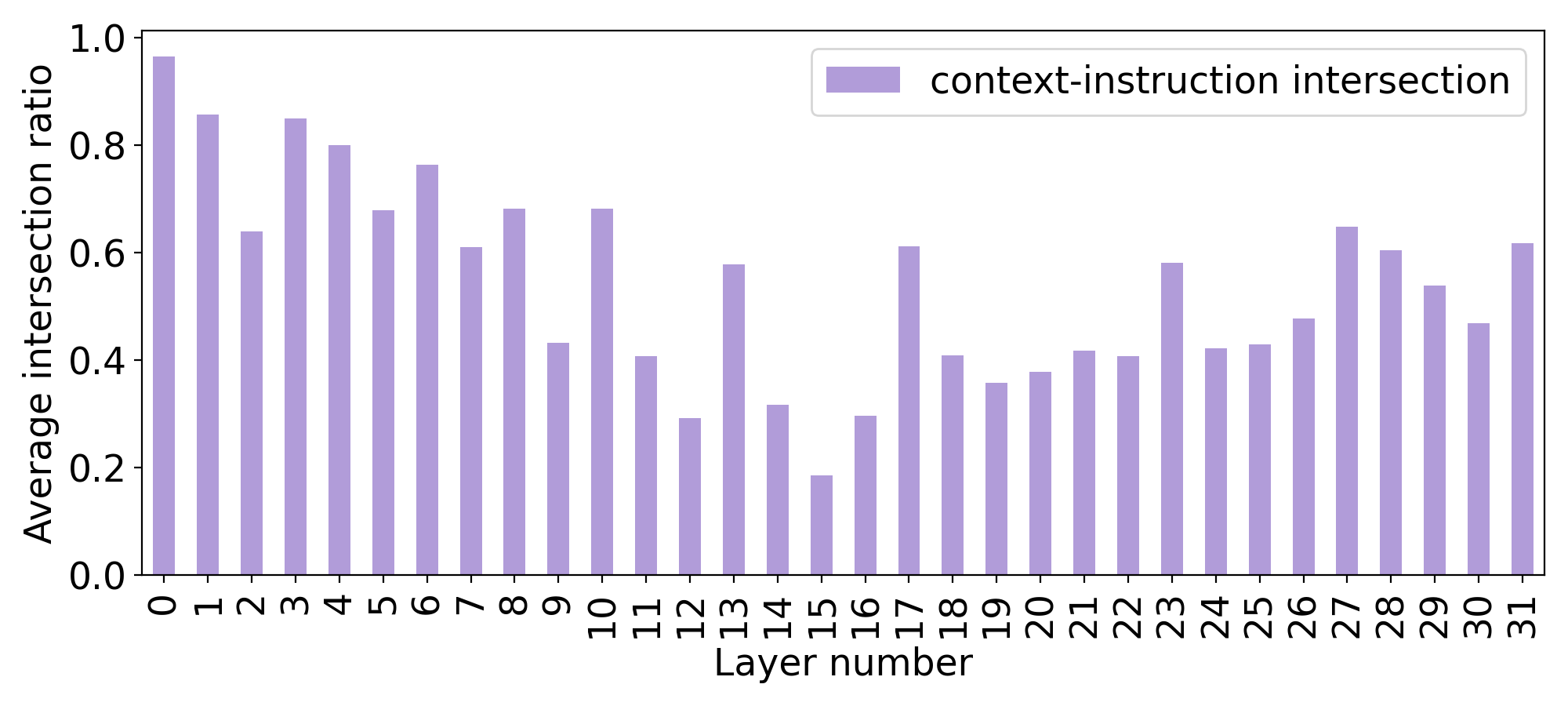} 
\caption{The difference between the top-$k$ hidden states selected by the instruction text and the document context with the $k$ set as $20$, conducted with Mistral 7B Instruct. Context-instruction intersection represents the overlap between the top-$k$ hidden states selected by the attention distribution from one piece of the context in the long document and the instruction text to a key-value cache.}
\label{fig:difference}
\end{figure}


The information neglect problem stems from the observation that the preserved states useful for language modeling are not necessarily the ones for a downstream task (e.g., answering a specific question). We demonstrate this by measuring the difference between the top-$k$ states selected by a document context compared to those selected by a specific instruction text (Figure~\ref{fig:overlap_method}). Specifically, we select one context and encode it to acquire a cache that could be evicted~(i.e, Context 1 in Figure~\ref{fig:overlap_method}). Then, we use another piece of context (i.e., Context 2 in Figure~\ref{fig:overlap_method}) and the instruction text, both with the same length, to evict the cache separately, retaining the top-$k$ most important hidden states. By computing the overlap of the differently evicted caches, we draw conclusions about the information neglect scenarios during the eviction-based language modeling process. More experimental setup for these experiments is shown in Appendix~\ref{app:setup_overlap}. We use the same setting to acquire the results in Figure~\ref{fig:intro_attention}.

We conduct this experiment on the full test set of the Qasper dataset~\citep{dasigi2021dataset}. We use the averaged attention score of all the tokens from one piece of text to the cache to select the most important states, which is further described in Section~\ref{sec:cse}.
As shown in Figure~\ref{fig:difference}, the hidden states focused on by the document context and the downstream instruction text are remarkably different, reflected by an intersection ratio lower than $0.2$ in the middle layers.

Supported by the above experiments, we claim that if only the attention distribution of the context chunk is used to select the key-value states relevant to language modeling, some information specifically related to the final instruction text will be discarded during the encoding of the document, possibly decrease the task performance.

A similar line of work that models long sequence with sliding-window-based methods \citep{xiao2023efficient, han2023lminfinite} also suffers from information neglect problems, where we provide detailed description in Appendix~\ref{app:sliding}.

Additionally, we conduct another set of experiments that demonstrate the performance degradation of the standard state eviction models compared to the standard models that use the full text as input. Results supporting the presence of information neglect problem are presented in Appendix~\ref{app:full_vs_stateeviction}.




\section{Methods}
\begin{figure}[t]
\centering
\includegraphics[width=0.95\columnwidth]{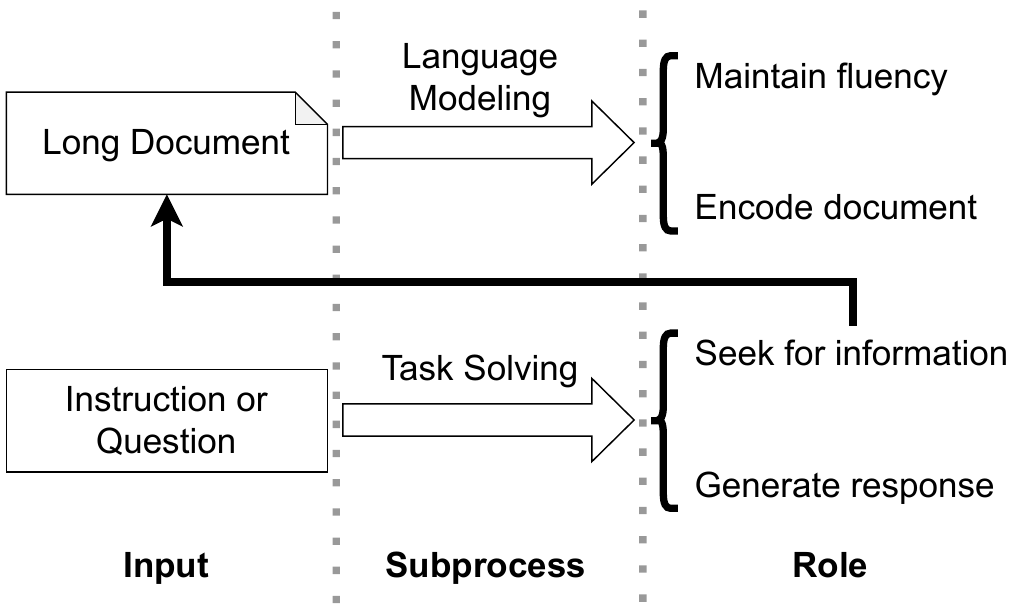} 
\caption{The illustration of our proposed different subprocesses for task-specific long sequence modeling. Each process serves as different roles.}
\label{fig:two_stage}
\end{figure}

To address the problem of information neglect, we propose to decompose the inference procedure of large language models into two different subprocesses: the language modeling process and the task solving process, shown in Figure~\ref{fig:two_stage}. For the language modeling process, we propose to use \textbf{chunked state eviction} methods to make the modeling process more efficient. For the task solving process, we propose \textbf{instruction-aware state eviction}, using the hidden states of the final instruction prompt as an additional instruction-aware query to extract and preserve the task-related information in a key-value cache. Then, we utilize this key-value cache to generate a task-specific response.

For downstream tasks with a long document input $D$ and a final instruction $I$ (a piece of text that prompt the model to conduct the downstream tasks), our proposed method generates a corresponding response $R$ according to $I$.

\subsection{Chunked State Eviction (CSE)}
\label{sec:cse}
In this section, we propose our standard state eviction method which chunks the input text during the language modeling process to enable the LLMs encoding the long document $D$ more efficiently. 

\begin{figure}[t]
\centering
\includegraphics[width=0.99\columnwidth]{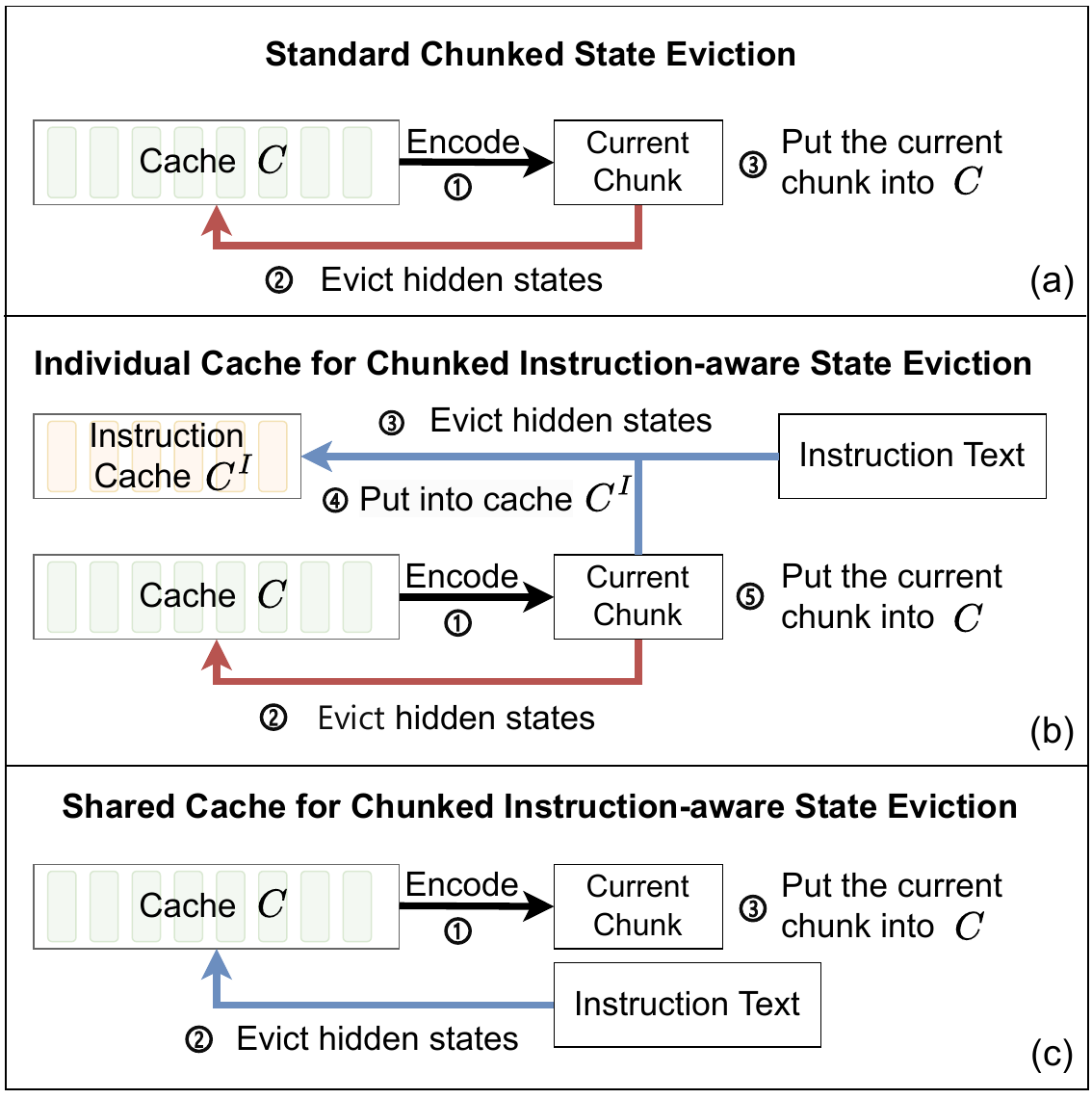} 
\caption{The illustration of different cache designs for our proposed Standard CSE and  \modelname{}.}
\label{fig:method}
\end{figure}

\paragraph{Overall process:} 
Given a document $D$, we divide it into chunks $D = \{s_1, s_2, \dots, s_n\}$, where $n$ denotes the number of chunks. Each chunk $s$ has a length of $l_s$ except for the final chunk $s_n$.
As illustrated in Figure~\ref{fig:method}(a), the Standard Chunked State Eviction (Standard CSE) process includes three steps: 1) given a cache $C$, we encode the current text chunk $s$ with an LLM;
2) evict the unimportant hidden states in $C$ according to the attention distribution from $s$ to $C$; 3) put all the new hidden states of $s$ into the cache. 
This iterative process starts with putting the first text chunk into the cache $C$ and ends when the document has been fully processed. After the whole encoding process, the final chunk (maybe shorter than the length of $l_s$) is put into the cache, which leads to possible information bias towards this chunk. To alleviate this bias, we use the instruction text as a new text chunk to evict the cache $C$ one more time. The resulting cache $C$ is then used to encode the instruction text and generate the final response.

\paragraph{State eviction based on chunked averaged attention score:} For state eviction, we use the attention score from all the tokens in the current text chunk $s$ to a state $c$ in the cache $C$ as a metric of the state's importance:
\begin{equation}
\text{Imp}(s, c) = \frac{1}{|s|} \sum_{t \in s} \frac{\exp\left(\frac{Q_t K_c^T}{\sqrt{d_k}}\right)}{\sum_{c' \in C} \exp\left(\frac{Q_t K_{c'}^T}{\sqrt{d_k}}\right)}
\end{equation}
where $\text{Imp}(s, c)$ represents the importance score of state $c$ with chunk $s$, $d_k$ is the dimensionality of the key vector, $ Q_t $ and $ K_c$ is the query vector of token $ t $ and the key vector of state $c$, respectively. 

We preserve the states with the $k$ highest importance scores while evicting the other states: 
\begin{equation}
   \mathbb{I}(x) = 
\begin{cases}
1, & \text{if } x \text{ in Set}_{\text{selection}} \\
0, & \text{otherwise}
\end{cases}
\label{eq1}
\end{equation}
\begin{equation}
\hat{C} = \{ c \in C \mid \mathbb{I}(c) = 1 \}
\label{eq2}
\end{equation}
where $\text{Set}_\text{selection}$ is the hidden states with the $k$ highest importance scores $\text{Imp}(s, c)$ from the current chunk $s$, and $\hat{C}$ represents the cache after the eviction.
We execute the eviction in a layer-wise manner, which means that the hidden states retained in different layers could belong to different tokens. This design allows more flexibility since different layers could be responsible for different functions and semantics. 
We choose to not apply a finer-grained head-wise eviction to our model since it performed worse in our initial experiments.




\subsection{Instruction-aware State Eviction} 



Next, we introduce chunked instruction-aware state eviction (\modelname{}) that aims to preserve information relevant to the task-solving process.
We propose two kinds of cache design to achieve this goal. First, we propose to maintain a separate individual instruction cache $C^I$ during the standard chunked state eviction process, which retains information related to the instruction text. Second, we propose a variant with a common shared cache for $C^I$ and $C$ to reduce the computational cost. 
Illustrations of the two proposed methods are shown in Figure~\ref{fig:method}.
\paragraph{Individual cache} We use an individual instruction cache $C^I$ to specifically store the hidden states related to the instruction text, in addition to $C$. Specifically, after the eviction on $C$, we conduct another eviction process on $C^I$ with the final instruction text, and then put the key-value states of the current text chunk $s$ into $C^I$. The eviction process is shown as follows:
\begin{equation}
   \mathbb{I}^I(x) = 
\begin{cases}
1, & \text{if } x \text{ in Set}^I_{\text{selection}} \\
0, & \text{otherwise}
\end{cases}
\label{equation:eq3}
\end{equation}
\begin{equation}
\hat{C^I} = \{ c^I \in C^I \mid \mathbb{I}^I(c^I) = 1 \}
\label{equation: eq4}
\end{equation}
where $\text{Set}^I_{\text{selection}}$ is the key-value cache states with $k$ highest importance scores of $\text{Imp}(I, c^I)$.

\paragraph{Shared cache} Using individual caches will double the memory usage for a fixed cache size. Guided by the persistence of importance hypothesis~\citep{liu2024scissorhands}, 
where the hidden states useful for maintaining the perplexity are attended by most of the following tokens, we hypothesize that the intersection between states selected by context and instruction texts, mentioned in Section~\ref{sec:information_neglect}, could be responsible for maintaining the perplexity.
Hence, we suppose that we could further reduce the memory cost of $C^I$ by sharing it with the language modeling process. 
Specifically, the top-$k$ state $\text{Set}_\text{selection}$  of the shared cache is determined based on the attention-based importance score $\text{Imp}(I, c)$, which measures the attention from the final instruction $I$ to a cache state $c$. 
Shown in Figure~\ref{fig:method}(c), we directly use this key-value cache evicted by $\text{Imp}(I, c)$ to encode the current chunk $s$.
The rest of the eviction process follows the same procedure as described in Eq. (\ref{eq1}) and (\ref{eq2}).

\subsection{Overall Process}
\label{sec:task_solving}

 
In this section, we summarize the overall process for applying CItruS to downstream tasks. 
As described in Section~\ref{sec:cse}, 
the model starts by iteratively encoding the chunked document $D$. 
Unlike the Standard CSE model, CItruS introduces the instruction text to evict either an individual or shared instruction-aware cache.
As mentioned, we use the instruction text to evict these caches again after processing the entire document, selecting the $k$ most important key-value states for each layer. We use these $k$ states to encode the final instruction and generate the response, thereby setting the size of each cache for all models to $k$ during this period\footnote{The cache size of our standard CSE and shared cache \modelname{} during the encoding is $l_s + k$
while the individual cache \modelname{} requires a cache size of $2\times(l_s + k)$.}.






\section{Experimental Setup}






\subsection{Tasks}

We compare the models using the following tasks. Detailed information about dataset statistics, prompts, and the divisions of document and instruction are provided in Appendices~\ref{app:dataset_statistic} and~\ref{app:prompt}.

\paragraph{Long document reading comprehension}
This task involves testing the ability of the models to answer a designated question based on a long document that exceeds the typical input length used during the pretraining of the large language models.
In this task, we use the datasets of Qasper~\citep{dasigi2021dataset}, MultifieldQA-en~\citep{bai2023longbench}, HotpotQA~\citep{yang2018hotpotqa}, and TriviaQA~\citep{joshi2017triviaqa}. We also include two other long few-shot tasks, Trec~\citep{li-roth-2002-learning} and SamSum~\citep{gliwa2019samsum}, which focus on classification and dialogue summarization, respectively. 
We follow \citet{bai2023longbench} to adapt these datasets into long-document tasks.
Instead of reporting the average scores in the main paper, we choose to report the average rank each model performs to avoid the variance differences among the datasets. Detailed results on each dataset is provided in Appendix~\ref{app:results_dataset}.

\paragraph{Long document knowledge retrieval} We use two tasks to test if the model could preserve the important information during the whole language modeling process: passkey retrieval\footnote{\url{https://huggingface.co/datasets/lvwerra/needle-llama3-16x524k}}~\citep{mohtashami2023landmark} and needle-in-a-haystack~\footnote{\url{https://github.com/gkamradt/LLMTest_NeedleInAHaystack}} tasks. The passkey retrieval task tests if the model can retrieve a single passkey (e.g., a five-digit number) inserted in a synthetic long document made up by repetitive simple sentences. We conduct this task on the documents with lengths up to 1 million tokens. The needle-in-a-haystack task replaces the passkey with a more general text fact and inserts them in real long documents. An example of the fact and the information of the documents can be found in Appendix~\ref{app:needle}. The maximum length of documents for needle-in-a-haystack is set to 12{,}000. We use accuracy in the passkey retrieval task and the ROUGE metric~\citep{lin-2004-rouge} for the needle-in-a-haystack task to award partial correctness. 
Additional experiments using BABILong~\citep{kuratov2024babilong}, a dataset design for the long-context needle-in-a-haystack task, are also conducted in Appendix~\ref{app:results_babilong}.

\paragraph{Long-range language modeling}
We report perplexity scores on long-range language modeling to estimate how well our models maintain fluency in generation~\citep{xiao2023efficient}. We used PG19~\citep{rae2019compressive} dataset for this task. 

\subsection{Baselines}
\label{sec:baselines}
\paragraph{Streaming LLM} always keeps the initial few tokens and uses a sliding window to model the long sequence~\citep{xiao2023efficient}. This model is known for its ability of modeling long sequences with lengths up to 4 million tokens.
\paragraph{TOVA} frames transformers as multi-state RNNs by using the attention distribution of the last token to identify which token should be evicted~\citep{oren2024transformers}. This model could be seen as a special case of our standard CSE model with the $l_s$ as 1. 
\paragraph{H2O} uses the accumulative attention score each token received to determine whether the token should be evicted~\citep{zhang2024h2o}.
\paragraph{RoCo} uses averaged attention probability from future tokens and determines the eviction scope by evaluating the standard variance of the attention weights one token receives~\citep{ren2024efficacy}.

\paragraph{}
\textsc{LongHead}~\citep{lu2024longheads} is another method that does not require further training. However, it requires large excessive memory cost (although could be offloaded to cpu memory, but that would cost more time) compared to our methods. Hence we choose to omit this model from our baselines to maintain a fair comparison.

Note that our proposed chunked instruction-aware state eviction is uncoupled with the eviction strategies used by the above models, hence it could be applied to all the above methods to achieve even better results. Due to the limitation of the computational cost, we only experiment the instruction-aware state eviction with our proposed chunked average attention score strategy and the accumulative attention score strategy used by H2O (denoted as H2O + Shared Cache) in our paper. All baselines are reimplemented with public repositories\footnote{\url{https://github.com/mit-han-lab/streaming-llm}}\footnote{\url{https://github.com/DRSY/EasyKV}}. For all baseline models, we apply the same encoding and generation process described in Section~\ref{sec:task_solving} for fair comparison.

\begin{table}[t]
  \centering
  \resizebox{\linewidth}{!}{
    \begin{tabular}{lrrrrrrrrrrrrrrrrrrrrrr}
    \toprule
\multirow{2}*{Settings} & \multicolumn{3}{c}{Mistral 7B Instruct} & \multicolumn{3}{c}{Llama 2 7B Chat} & \multicolumn{3}{c}{Llama 2 13B Chat}  \\ 
      \cmidrule(r{4pt}){2-4} \cmidrule(r{4pt}){5-7} \cmidrule(r{4pt}){8-10}
     & 0-4k & 4k-8k & 8k+ & 0-4k & 4-8k  & 8k+ & 0-4k & 4-8k & 8k+ \\
    \midrule

     Streaming LLM  &  2.83 & 3.17 & 3.50 & 2.50 & 3.00 & 4.17 & 1.67 & 3.17 & 3.83  \\
     TOVA  &   2.67 & 3.00 & 2.67 & 3.67 & 4.00 & 3.50 & 3.83 & 4.00 & 4.33   \\
     RoCo  &   3.67 & 2.67 & 2.83 & 3.00 & 3.17 & 2.00 & 4.00 & 1.33 & 2.33  \\
     H2O  & 4.00 & 3.50 & 4.17 & 4.17 & 2.50 & 2.67 & 3.33 & 3.50 & 4.83 \\
    \midrule
      Standard CSE &  3.33 & 3.67 & 3.00 & 3.67 & 4.17 & 4.17 & 5.00 & 3.50 & 2.00 \\
     \quad $+$ Individual Cache  & \textbf{7.17} & \textbf{8.00} & \textbf{7.00} & 6.50 & 7.00 & 6.67 & 6.50 & 7.33 & 6.33  \\
     \quad $+$ Shared Cache & 6.50 & 6.67 & 7.00 & \textbf{6.83} & \textbf{7.33} & \textbf{7.33} & \textbf{6.50} & \textbf{7.33} & \textbf{6.33} \\
     H2O $+$ Shared Cache & 5.17 & 5.17 & 5.50 & 5.33 & 4.83 & 5.50 & 4.67 & 5.17 & 5.83 \\

    \bottomrule
    \end{tabular}
    }

\caption{The averaged reversed rank results among all the 8 models on six different reading comprehension tasks, where 8 is the highest score and 1 is the lowest score. Results are presented by grouping text with lengths of 0-4k, 4k-8k, and 8k$+$. Best results are bolded.}
    \label{tab:benchmark_rank}
\end{table}

\begin{figure*}[t]
\centering
\includegraphics[width=1.99\columnwidth]{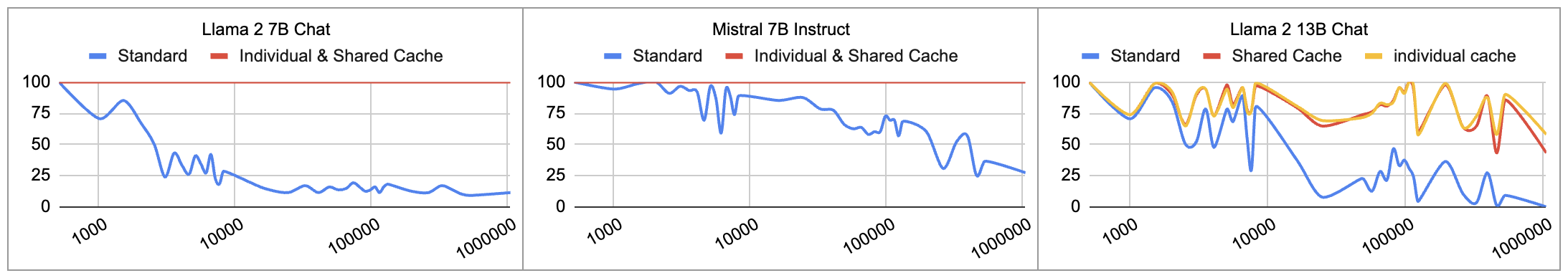}
\caption{The results of the passkey retrieval task with Llama 2 7B Chat, Mistral 7B Instruct, and Llama 2 13B Chat. }
\label{fig:passkey}
\end{figure*}

\subsection{Hyperparameters}
We applied the position shift mechanism leveraged by~\citet{xiao2023efficient}, which always use the same positional embeddings for the caches containing different hidden states, to make the models process long documents better. We also apply this technique to all the baselines to enhance their ability of processing long sequences.
We use the Llama 2 Chat model~\citep{touvron2023llama} with 7 billion and 13 billion parameters and the 7 billion parameter Mistral Instruct model~\citep{jiang2023mistral} as the backbone models. 
Additionally, we include experiments using Llama 3 8B Instruct model, shown in Appendix~\ref{app:results_llama3}.
$k$ is set as $768$ and $l_s$ is set as $256$, resulting a cache size of $1{,}024$ during modeling the document. This setting is also applied to all the baseline models. We apply 8 bit quantization on the 13 billion parameter model. Results are inferred on one A100 80G GPU. All the hyperparameters are selected using the validation sets.


\begin{table}[t]
  \centering
  \resizebox{0.99\linewidth}{!}{
    \begin{tabular}{llrrrrrrrrr}
    \toprule
      \multirow{2}*{Settings} & \multicolumn{3}{c}{Llama 2 7B Chat} & \multicolumn{3}{c}{Mistral 7B Instruct} \\ 
    \cmidrule(r{4pt}){2-4}\cmidrule(r{4pt}){5-7}
    ~ & R-1 & R-2 & R-L & R-1 & R-2 & R-L \\
        \midrule
    \multicolumn{7}{c}{$l_s = 256, k = 768$} \\
    \midrule
    Standard CSE &   19.87 &   5.74 & 17.52  & 15.17 & 6.34 &  13.94 \\
     \quad $+$ Individual Cache  &  \textbf{24.72} &  \textbf{7.87} & \textbf{24.53} & 59.05 & 51.22 & 59.10  \\
     \quad $+$ Shared Cache & 23.73 & 7.46& 23.44 & \textbf{63.47} & \textbf{55.33} & \textbf{63.43} \\

    \midrule
    \multicolumn{7}{c}{$l_s = 1{,}024, k = 1{,}024$} \\
    \midrule
    Standard CSE & 18.86 & 7.52 & 18.04  & 30.21 & 14.18 & 29.23  \\
     \quad $+$ Individual Cache  & \textbf{33.28} & 17.52 & \textbf{32.76} & 56.12 & 51.20 & 56.08  \\
     \quad $+$ Shared Cache & 31.95 & \textbf{18.41} & 31.47 & \textbf{57.15} & \textbf{51.60} & \textbf{56.97} \\

    \bottomrule
    \end{tabular}
    }

\caption{Results of the needle-in-a-haystack task. Best results are bolded. R-1, R-2, and R-L represent ROUGE-1, ROUGE-2, and ROUGE-L, respectively.}
    \label{tab:needle_main}
\end{table}

\begin{figure}[t]
\centering
\includegraphics[width=0.99\columnwidth]{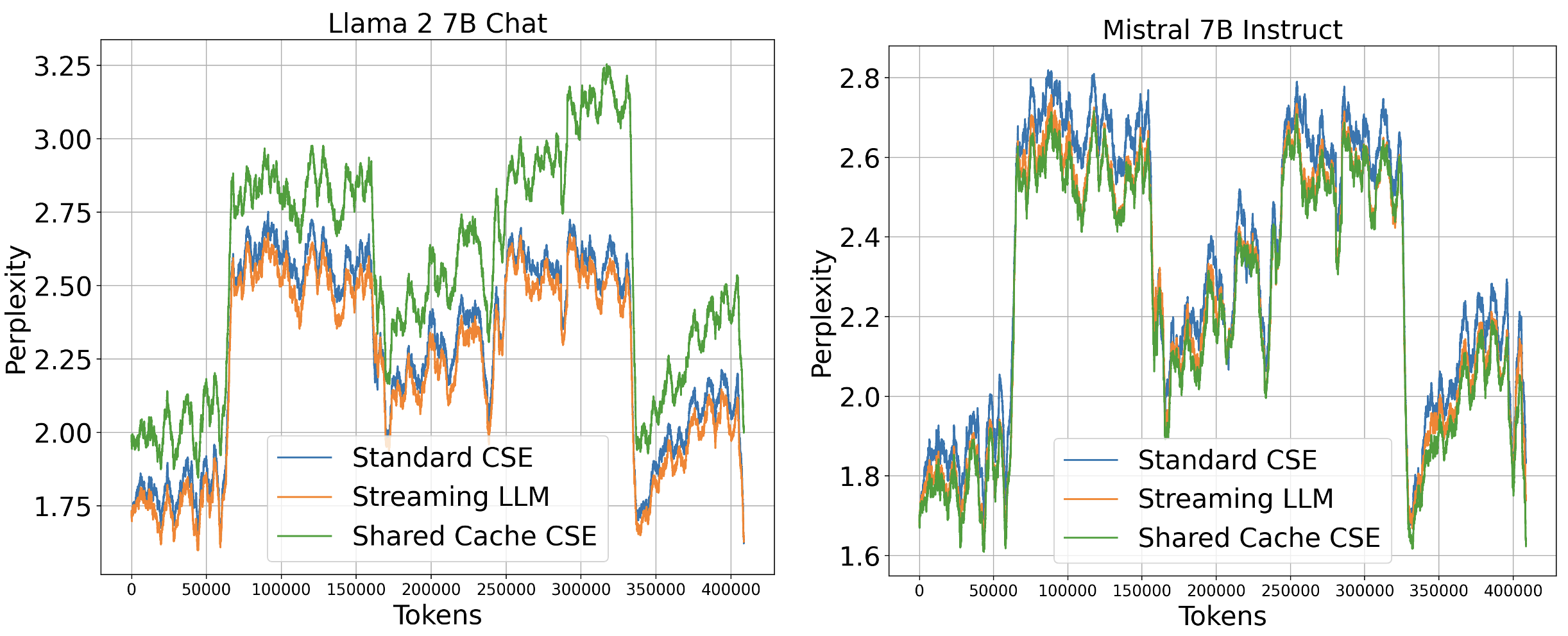} 
\caption{The language modeling results on the Llama 2 7B chat and Mistral 7B Instruct model. The line chart is smoothed with a window size of $4096$ for clarity.}
\label{fig:ppl}
\end{figure}

\section{Results}

\subsection{Long document reading comprehension}
The results of the long document reading comprehension tasks aggregated over six datasets are shown in Table~\ref{tab:benchmark_rank}, while the dataset-specific results are shown in Appendix~\ref{app:results_dataset}.
First, our Standard CSE method achieves performance comparable to all the baselines, demonstrating the effectiveness of our basic framework.
Both variants of \modelname{} consistently outperform all baselines and Standard CSE. 
As mentioned in Section~\ref{sec:baselines}, our method could also be applied on different eviction policies.
Hence, we further included a variant of the H2O model (H2O + Shared Cache) and show that it achieves better performance over the H2O model in all cases.

We find models with a shared cache achieve the same level of performance as their corresponding model with separate caches. This suggests that the overlapping tokens between the context and the instruction text might be sufficient to support language modeling, while the shared cache also maintains the information useful for the downstream tasks. We will further discuss this in Section~\ref{sec:language_modeling}. 


\subsection{Long document knowledge retrieval}
The main results of long document knowledge retrieval are shown in Figure~\ref{fig:passkey} and Table~\ref{tab:needle_main}. 
Our proposed \modelname{} 
retrieves all the passkeys
using Llama 2 7B and Mistral 7B while still outperforming the Standard CSE for Llama 2 13B\footnote{We omit 5 outlier data points from all the 38 data points for Llama 2 13B Chat in the passkey retrieval tasks where all the models performs with an accuracy of 0\%.}, which shows the superiority of \modelname{} for long document knowledge retrieval.
For the needle-in-a-haystack task, our method outperforms the standard state eviction methods across different large language models and lengths.


\subsection{Long-range language modeling}
\label{sec:language_modeling}
We compare our model with the long-range language modeling model, Streaming LLM. Specifically, we evaluate the standard CSE as well as the shared cache version of our proposed \modelname{}. For \modelname{} with a shared cache, we randomly sample 10 different instructions including different questions from Qasper and HotpotQA dataset. We show the results using one instruction here and append the rest of the results in the Appendix~\ref{app:other_instruction_ppl}. Results in Figure~\ref{fig:ppl} show that our standard CSE could maintain the perplexity when processing long sequences as low as the Streaming LLM. 
Meanwhile, although showing a slight increase in perplexity with the Llama 2 7B Chat model, CSE with a shared cache achieves consistent perplexity results without exploding as described by \citet{xiao2023efficient}.
This shows that introducing the instruction text as the query to evict hidden states would not affect the text perplexity of the large language models.
A more detailed discussion about the roles of the standard cache and the instruction-aware cache in our model is provided in Appendix~\ref{app:discussion}.

\subsection{Analysis}
In this section, we provide analyses on the hyper-parameters of our model, the effect of chunk size, and the position bias in the knowledge retrieval tasks. We also provide an analysis on the effect of the initial tokens in Appendix~\ref{app:start_size}.  We report the averaged results in this section since all the models perform similarly in the analyses across all the datasets. The full results are shown in Appendix~\ref{app:results_dataset}.

\begin{table}[t]
  \centering
  \resizebox{0.99\linewidth}{!}{
    \begin{tabular}{lllrrrrrrrrr}
    \toprule
     \multirow{2}*{Param.} & \multirow{2}*{Settings}  & \multicolumn{3}{c}{Llama 2 7B Chat} & \multicolumn{3}{c}{Mistral 7B Instruct} \\ 
    \cmidrule(r{4pt}){3-5}\cmidrule(r{4pt}){6-8}
    ~ & ~ & 0-4k & 4-8k & 8k+ & 0-4k & 4-8k & 8k+ \\
    \midrule
    \multirow{3}*{\makecell[l]{$l_s = 256$\\ $k = 768$}}& Standard CSE &  36.72 & 37.07 & 38.36  & 34.52 & 30.57 & 20.92 \\
     &\quad $+$ Individual Cache  & \textbf{43.45} & 43.26 & 45.93 & \textbf{45.15} & \textbf{45.11} &\textbf{41.55} \\
     &\quad $+$ Shared Cache & 43.22 & \textbf{44.07} & \textbf{46.37}  & 43.96 & 41.56 & 36.61  \\

    \midrule
    \multirow{3}*{\makecell[l]{$l_s = 512$\\ $k = 512$}}& Standard CSE & 36.92 & 34.51 & 35.07 & 33.98 & 28.36 & 21.92  \\
    & \quad $+$ Individual Cache  & 41.02 & 41.52 & 41.81 & \textbf{45.13} & \textbf{43.38} & \textbf{41.66} \\
    & \quad $+$ Shared Cache & \textbf{41.17} & \textbf{41.79} &\textbf{43.57} & 44.65 & 40.06 & 35.51 \\
    \midrule
    \multirow{3}*{\makecell[l]{$l_s = 768$\\ $k = 256$}}& Standard CSE & 32.59 & 31.04 & 29.57 & 30.60 & 23.19 & 21.44   \\
    & \quad $+$ Individual Cache  & 34.73 & 33.79 & 33.88 & \textbf{40.82} & \textbf{36.67} & \textbf{33.04} \\
    & \quad $+$ Shared Cache &  \textbf{36.12} & \textbf{35.67} & \textbf{34.61} & 38.89 & 32.50 & 28.77  \\
    \bottomrule
    \end{tabular}
    }

\caption{Results of the hyperparameters under the same memory budget. Best results are bolded. ``Param.'' stands for hyperparameters.}
    \label{tab:hyperparameter}
\end{table}

\begin{table}[t]
  \centering
  \resizebox{0.99\linewidth}{!}{
    \begin{tabular}{lllrrrrrrrrr}
    \toprule
     \multirow{2}*{Param.} & \multirow{2}*{Settings}  & \multicolumn{3}{c}{Llama 2 7B Chat} & \multicolumn{3}{c}{Mistral 7B Instruct} \\ 
    \cmidrule(r{4pt}){3-5}\cmidrule(r{4pt}){6-8}
    ~ & ~ & 0-4k & 4-8k & 8k+ & 0-4k & 4-8k & 8k+ \\

    \midrule
    \multirow{3}*{\makecell[l]{$l_s = 64$\\$k=768$}}& Standard CSE &    36.70 & 35.83 & 38.12 &  32.37 & 28.46 & 20.72 \\
    & \quad $+$ Individual Cache  &  \textbf{43.78} & \textbf{43.86} &\textbf{ 47.48} & \textbf{45.88} &\textbf{ 44.01} & \textbf{40.20}   \\
    & \quad $+$ Shared Cache &  43.09 & 42.82 & 43.24  & 43.81 & 38.89 & 34.36 \\
    
    \midrule
    \multirow{3}*{\makecell[l]{$l_s = 256$\\$k=768$}}& Standard CSE & 36.72 & 37.07 & 38.36 & 34.52 & 30.57 & 20.92  \\
    & \quad $+$ Individual Cache  &  \textbf{43.45} & 43.26 & 45.93 & \textbf{45.15} & \textbf{45.11} & \textbf{41.55} \\
    & \quad $+$ Shared Cache & 43.22 & \textbf{44.07} & \textbf{46.37} & 43.96&	41.56	& 36.61 \\
    \midrule
    \multirow{3}*{\makecell[l]{$l_s = 512$\\$k=768$}}& Standard CSE &  38.17 & 37.49 & 37.71 & 31.99 & 27.90 & 24.39 \\
    & \quad $+$ Individual Cache  & \textbf{43.04} & \textbf{43.18} & \textbf{46.71} &\textbf{ 42.79} & \textbf{42.09} & \textbf{40.58} \\
    & \quad $+$ Shared Cache &  42.60 & 42.89 & 46.25 & 42.43 & 40.05 & 35.16 \\
    \midrule
    \multirow{3}*{\makecell[l]{$l_s = 768$\\$k=768$}}& Standard CSE &   39.41 & 38.25 & 38.39 & 33.35 & 25.74 & 20.51 \\
    & \quad $+$ Individual Cache  & \textbf{42.27} & \textbf{43.45} & 42.26 & 42.31 & \textbf{39.76} & \textbf{37.37}  \\
    & \quad $+$ Shared Cache &  42.09 & 42.61 & \textbf{43.76} & \textbf{42.76}& 38.72& 33.56 \\
    \bottomrule
    \end{tabular}
    }

\caption{Results of the different chunk size $l_s$. Best results are bolded. ``Param.'' stands for hyperparameters. }
    \label{tab:chunk_size}
\end{table}

\subsubsection{Hyperparameter analysis}
Given a fixed memory budget, there is a trade off between $l_s$ and $k$. A larger $k$ can preserve more information, potentially leading to a better performance, and $l_s$ affects the encoding efficiency. 
In this section, we probe our model by adjusting different hyperparameters to demonstrate that our proposed \modelname{} is insensitive to them. 

Table~\ref{tab:hyperparameter} shows that with a fixed budget, \modelname{} consistently outperforms the Standard CSE models, showing that our method is not sensitive to the choices of $k$ and $l_s$, and the instruction-aware cache methods are the best when considering both the efficiency and the down-stream task performance.


\subsubsection{Analysis of the chunk size}
We provide a comparison of models using chunk sizes ranging from $64$ to $768$.
The inference time of each model decreases linearly as $l_s$ increases. 
 
As shown in Table~\ref{tab:chunk_size}, the performance fluctuation when using different chunk sizes is very limited, while the efficiency is significantly improved. Our \modelname{} model extends the chunk size beyond that of previous methods and demonstrates a substantial improvement in efficiency for conducting long-sequence downstream tasks.


\begin{figure}[t]
\centering
\includegraphics[width=0.99\columnwidth]{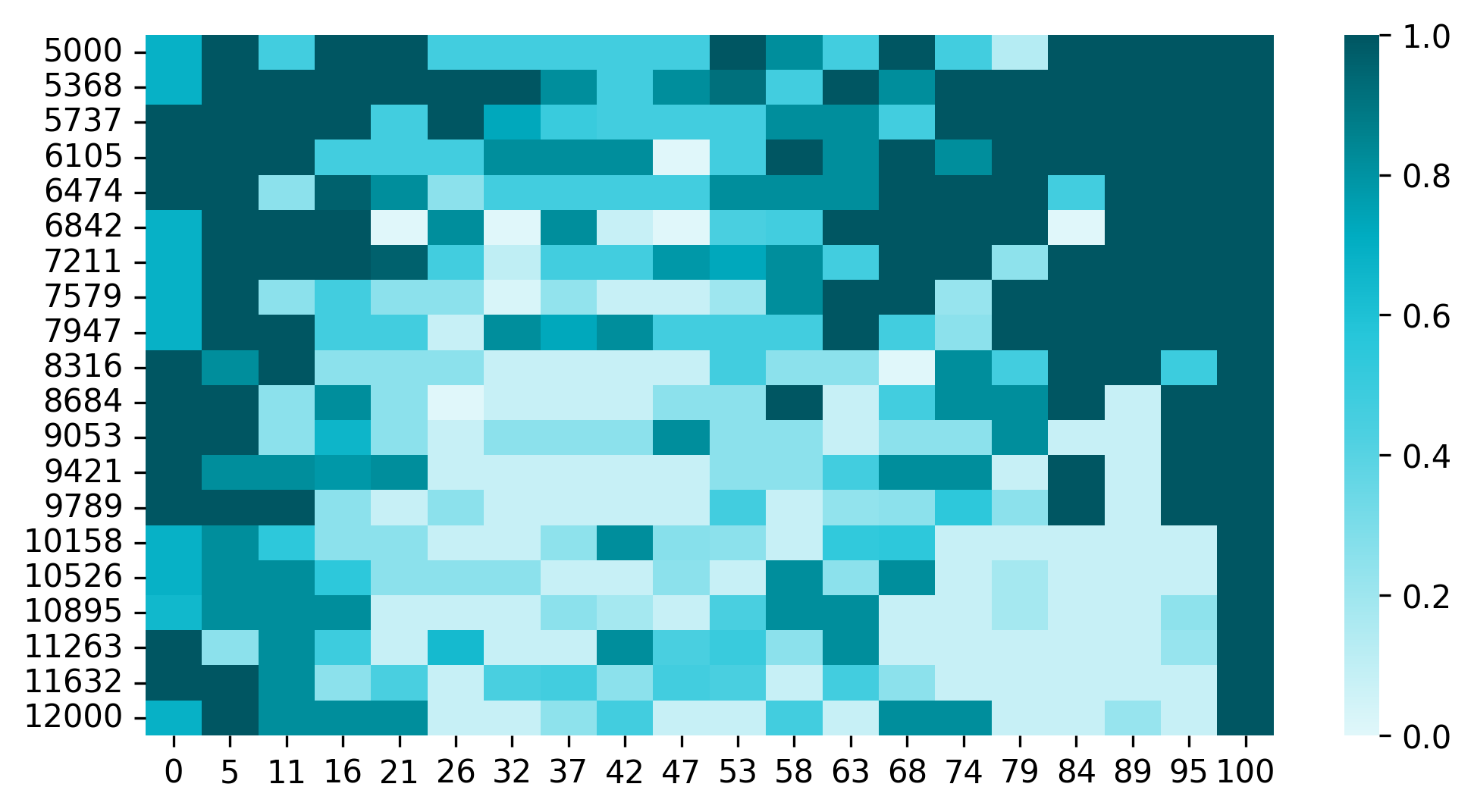}
\caption{The position-wise results from CItruS with shared cache ($l_s=1{,}024, k=1{,}024$) on needle-in-the-haystack using Mistral 7B Instruct. The x-axis represents the position where the needle is inserted, while the y-axis represents the length of the documents. The color of the grid represents the ROUGE-1 score.}
\label{fig:needle_position}
\end{figure}

\subsubsection{Position bias in knowledge retrieval}
\citet{liu2023lost} propose that large language models tend to pay less attention to the middle parts of long documents. In this section, we test our model to determine if this issue persists with our proposed instruction-aware cache method.


We use the needle-in-a-haystack task as the basic task and evaluate the ROUGE results when the fact is inserted at different positions in the document. As shown in Figure~\ref{fig:needle_position}, we demonstrate that the CItruS model still prefers to attend to the information at the beginning and the end, leaving future work to address this lost-in-the-middle issue in eviction-based long-sequence methods.

\section{Conclusion}
We have proposed \modelname{}, an inference-time state eviction method for large language models (LLMs) that improves their performance on long sequence downstream tasks. It features a large chunked sequence processing procedure and an instruction-aware cache that helps with solving downstream tasks. Experiments on long document reading comprehension, knowledge retrieval, and language modeling show the utility of our method compared to strong baselines. 

Our work demonstrates the possibility of generalizing standard LLMs trained on text constrained to certain lengths to processing longer sequences without any parameter adjustments. 
Our evaluation mainly focuses on retrieving task-related information from a long document. Future work may consider extending more high-level abilities~(e.g., multi-hop and compositional reasoning) to the long sequence regime. Moreover, trainable components can be further introduced to facilitate this process.

\section*{Acknowledgements}
The authors thank all the reviewers for their suggestions and comments. This work is supported by National Natural Science Foundation of China (No. U21B2009) and McGill Science Undergraduate Research Award (SURA). Jackie Chi Kit Cheung is supported by Canada CIFAR AI Chair program. The authors also acknowledge the material support of NVIDIA in the form of computational resources.

\section*{Limitations}
We only tested our methods with Llama 2 and Mistral models, leaving performance on other datasets to be evaluated.
The instruction-aware cache is only applied to our Standard CSE and the H2O models, it could be further applied to models using other state eviction policies to possibly further enhance the performance.
Our work only uses one instruction for each task to conduct all the experiments. It would be interesting to show whether better instruction texts exist that are specifically designed for conducting long sequence down-stream tasks.
Future work might consider optimizing the query, or even use soft prompt optimization technique to select the hidden states.

\section*{Ethical Considerations}
The associated risks with this work include using a model trained on vast amounts of text, which likely contains gender, racial, and cultural bias. Another concern is the potential misuse of the model for generating misleading or harmful content when applying our method to generate text. Meanwhile, cache-based methods could be more effective for malicious applications like jailbreaking or revealing private information, since it breaks the standard usage of the hidden states in large language models.

\clearpage
\bibliography{anthology,custom}

\clearpage
\appendix

\section{Details for the Intersection Probing Experiments}
\label{app:setup_overlap}

The goal of the intersection probing experiment is to determine whether the document context selects a different set of top-$k$ states with the highest attention scores  within the cache compared to the instruction text. This difference could lead to the document context overlooking crucial information required by the final instruction.

For this purpose, we use all the 416 documents in the test split of the Qasper dataset \citep{dasigi2021dataset}. For each document, we randomly select a chunk, referred to as Context 1, consisting of 200 tokens from the first $\frac{1}{4}$ document to simulate the cache $C$ during the eviction process. If the first $\frac{1}{4}$ document contains fewer than 200 tokens, we use the entire first $\frac{1}{4}$ as Context 1. Then, we randomly select a second chunk, referred to as Context 2, from the final $\frac{1}{4}$ document to ensure sufficient distance between Context 1 and Context 2, avoiding recency bias and placing Context 2 close to the final instruction text. To ensure a fair comparison, we also make sure that the length of Context 2 is the same as that of the instruction text for each document. 

We send the concatenation of Context 1 and Context 2 to the Mistral 7B Instruct model to obtain the simulated cache $C$, which consists of all the key-value states of Context 1. We could also acquire the attention distribution from Context 2 to Context 1 through this step. At each model layer $l$, we define the importance of the $j$th state in Context 1 as the average of $a^l_{ij}$, the attention score from each position $i$ in Context 2 to the $j$th state in Context 1. We keep the top-$k$ states in Context 1 with the highest average attention scores as $\text{Set}_{\text{selection}}$, and compute the final evicted
cache $\hat{C}_{\text{context 2}}$ following Equ. (\ref{eq1}) and (\ref{eq2}). Similarly, we use the same model to encode the concatenation of Context 1 and the instruction text to get the attention distribution from the instruction text to Context 1, and follow the same steps as described above to obtain the final evicted cache $\hat{C}_{\text{instruction}}$ from the instruction text. In this experiment, we set $k$ to $20$, which is $\frac{1}{10}$ of length of the first context.

We compute the intersection ratio between the $\hat{C}_{\text{context 2}}$ and $\hat{C}_{\text{instruction}}$ as $\frac{|\hat{C}_{\text{context 2}} \cap \hat{C}_{\text{instruction}}|}{|\hat{C}_{\text{instruction}}|}$, and average the intersection ratio over all the 416 documents for each layer. As shown in Figure~\ref{fig:difference}, the intersection ratio is particularly low in the middle layers of the model, supporting our hypothesis that the document context neglects a significant amount of information considered important by the final instruction. This discrepancy may be attributed to the remarkably different semantics of the instruction text and the document context, despite their close proximity.

\begin{table*}[t]
  \centering
  \resizebox{0.80\linewidth}{!}{
    \begin{tabular}{llrrrrrrrrrrrrrrrrrrrrrr}
    \toprule
\multirow{2}*{Settings} & \multicolumn{3}{c}{Llama 2 7B Chat} & \multicolumn{3}{c}{Llama 2 13B Chat} & \multicolumn{3}{c}{Mistral 7B Instruct}  \\ 
      \cmidrule(r{4pt}){2-4} \cmidrule(r{4pt}){5-7} \cmidrule(r{4pt}){8-10}
     & 0-4k & 4k-8k & 8k+ & 0-4k & 4-8k  & 8k+ & 0-4k & 4-8k & 8k+ \\
    \midrule

     Streaming LLM  &   33.78 & 34.92 & 37.11 & 37.39 & 37.95 & 36.54 & 34.26 & 31.00 & 27.21  \\
     TOVA  &  35.60 & 33.98 & 35.80 & 40.71 & 37.59 & 35.52 & 31.67 & 27.54 & 22.17  \\
     RoCo  &   32.46 & 25.70 & 20.64 & 40.30 & 31.02 & 25.89 & 32.67 & 25.28 & 19.83 \\
     H2O  &  34.47 & 29.54 & 27.26 & 38.68 & 35.24 & 35.96 & 34.60 & 25.37 & 23.08 \\
    \midrule
      Standard CSE &  36.72 & 37.07 & 38.36 & 43.68 & 39.18 & 30.74 & 34.52 & 30.57 & 20.92 \\
     \quad $+$ Individual Cache  & \textbf{43.45} & 43.26 & 45.93 & 46.43 & 46.61 & 40.80 & \textbf{45.15}& \textbf{45.11} & \textbf{41.55} \\
     \quad $+$ Shared Cache & 43.22 & \textbf{44.07} & \textbf{46.37} & \textbf{46.66} & \textbf{46.91} & 41.53 & 43.96 & 41.56 & 36.61 \\
     H2O $+$ Shared Cache & 38.26 & 39.19 & 40.27 & 42.00 & 41.54 & \textbf{42.30} & 40.28 & 36.23 & 32.45 \\
    \bottomrule
    \end{tabular}
    }

\caption{The averaged results on six different long sequence tasks. Results are separately presented by grouping text with different source lengths. Best results are bolded.}
    \label{tab:benchmark_avg}
\end{table*}

\begin{table*}[t]
  \centering
  \resizebox{0.99\linewidth}{!}{
    \begin{tabular}{llrrrrrrrrrrrrrrrrrrrrrr}
    \toprule
      \multirow{2}*{Models} & \multirow{2}*{Settings} & \multicolumn{3}{c}{Qasper} & \multicolumn{3}{c}{MultifieldQA} & \multicolumn{3}{c}{HotpotQA} & \multicolumn{3}{c}{Trec}  &\multicolumn{3}{c}{TriviaQA} & \multicolumn{3}{c}{SamSum}   \\ 
      \cmidrule(r{4pt}){3-5} \cmidrule(r{4pt}){6-8} \cmidrule(r{4pt}){9-11}\cmidrule(r{4pt}){12-14}\cmidrule(r{4pt}){15-17} \cmidrule(r{4pt}){18-20}
       & & 0-4k & 4k-8k & 8k+ & 0-4k & 4-8k  & 8k+ & 0-4k & 4-8k & 8k+ & 0-4k & 4-8k & 8k+ & 0-4k & 4-8k & 8k+ & 0-4k &  4-8k & 8k+ \\
    \midrule
        \multirow{8}*{\makecell[l]{Llama 2\\ 7B Chat}}
    & Streaming LLM &    8.36 & 10.54 & 27.77 & 23.51 & 22.07 & 17.34 & 25.56 & 23.81 & 26.13 & 47.00 & 52.00 & 40.00 & 61.96 & 66.64 & 71.37 & 36.31 & 34.47 & 40.07 \\
    & TOVA  &  9.81 & 13.01 & 25.00 & 22.44 & 23.99 & 16.06 & 35.16 & 29.15 & 29.66 & 50.00 & 57.00 & 50.00 & 63.95 & 53.78 & 63.03 & 32.23 & 26.94 & 31.06  \\
    & RoCo  &   10.82 & 15.71 & 7.89 & 27.39 & 15.74 & 12.43 & 30.99 & 27.74 & 20.48 & 49.00 & 59.00 & 56.00 & 53.20 & 28.20 & 21.53 & 23.35 & 7.79 & 5.53  \\
    & H2O   &  9.31 & 12.61 & 14.54 & 31.08 & 21.66 & 23.52 & 46.92 & 35.22 & 32.74 & 50.00 & 54.00 & 43.00 & 62.45 & 48.06 & 45.76 & 7.06 & 5.67 & 4.00  \\
    & Standard CSE &    8.43 & 14.84 & 27.08 & 23.19 & 22.67 & 15.06 & 33.30 & 24.68 & 35.27 & 49.00 & 55.00 & 52.00 & 70.10 & 75.05 & 71.99 & 36.27 & 30.17 & 28.77 \\
    & \quad $+$ Individual Cache & 20.71 & 16.80 & 28.77 & 36.65 & 28.69 & 24.67 & 43.04 & 38.95 & 41.75 & 55.00 & 62.00 & 65.00 & 67.44 & 80.30 & 82.10 & 37.87 & 32.84 & 33.26 \\
    & \quad $+$ Shared Cache &  20.78 & 17.48 & 29.02 & 38.43 & 30.81 & 24.90 & 43.21 & 40.59 & 42.88 & 56.00 & 63.00 & 63.00 & 64.14 & 80.05 & 83.50 & 36.74 & 32.48 & 34.93  \\
    & H2O $+$ Shared Cache &   16.91 & 14.87 & 33.52 & 40.59 & 31.80 & 25.78 & 44.00 & 38.32 & 36.07 & 45.00 & 53.00 & 49.00 & 75.27 & 73.26 & 66.71 & 7.76 & 23.88 & 30.55 \\

    \midrule
    \multirow{8}*{\makecell[l]{Llama 2\\ 13B Chat}}
    & Streaming LLM  &   12.48 & 12.86 & 19.81 & 24.36 & 24.61 & 14.22 & 28.66 & 30.12 & 31.88 & 52.00 & 58.00 & 44.00 & 76.09 & 77.91 & 75.22 & 30.76 & 24.19 & 34.12  \\
    & TOVA  &   17.18 & 13.76 & 23.20 & 26.75 & 24.34 & 14.23 & 40.79 & 29.56 & 36.59 & 56.00 & 59.00 & 49.00 & 80.05 & 84.22 & 76.43 & 23.51 & 14.66 & 13.69 \\
    & RoCo  &   16.09 & 12.86 & 24.41 & 26.85 & 17.48 & 10.08 & 39.09 & 26.26 & 15.84 & 57.00 & 58.00 & 45.00 & 82.41 & 69.71 & 59.27 & 20.34 & 1.79 & 0.72  \\
    & H2O  &  15.72 & 12.92 & 27.03 & 30.28 & 26.01 & 22.00 & 35.36 & 35.95 & 30.66 & 56.00 & 53.00 & 52.00 & 80.93 & 78.37 & 77.33 & 13.77 & 5.20 & 6.73  \\
    & Standard CSE   &   17.92 & 15.88 & 4.91 & 27.75 & 22.52 & 9.09 & 44.46 & 29.74 & 30.65 & 55.00 & 51.00 & 42.00 & 80.68 & 83.61 & 70.48 & 36.24 & 32.34 & 27.29  \\
    & \quad $+$ Individual Cache &   16.67 & 25.07 & 9.06 & 39.58 & 34.79 & 15.19 & 43.27 & 37.91 & 40.38 & 57.00 & 60.00 & 58.00 & 84.70 & 86.32 & 86.52 & 37.36 & 35.57 & 35.66 \\
    & \quad $+$ Shared Cache  & 19.52 & 25.75 & 15.47 & 38.22 & 34.36 & 22.34 & 45.78 & 37.83 & 40.54 & 55.00 & 63.00 & 56.00 & 85.30 & 87.79 & 83.89 & 36.11 & 32.70 & 30.96 \\
    & H2O $+$ Shared Cache &22.02 & 23.99 & 28.04 & 35.96 & 32.21 & 43.04 & 34.49 & 35.38 & 39.70 & 54.00 & 58.00 & 49.00 & 84.80 & 89.32 & 82.36 & 20.71 & 10.32 & 11.67 \\
            \midrule
    \multirow{8}*{\makecell[l]{Mistral\\ 7B Instruct}}
    & Streaming LLM &   26.90 & 20.21 & 13.59 & 38.51 & 27.72 & 17.17 & 29.71 & 24.94 & 27.42 & 48.00 & 55.00 & 44.00 & 49.06 & 45.27 & 49.39 & 13.39 & 12.84 & 11.70 \\
    & TOVA  &  27.82 & 21.95 & 13.86 & 38.70 & 28.07 & 18.04 & 32.68 & 25.01 & 23.52 & 47.00 & 54.00 & 37.00 & 36.96 & 27.96 & 28.93 & 6.84 & 8.22 & 11.68  \\
    & RoCo  & 28.35 & 26.06 & 18.43 & 45.18 & 27.45 & 17.84 & 47.24 & 35.26 & 26.77 & 45.00 & 48.00 & 44.00 & 22.74 & 8.32 & 6.86 & 7.53 & 6.57 & 5.05  \\
    & H2O  &   27.02 & 22.67 & 14.60 & 48.68 & 32.12 & 31.21 & 49.16 & 35.36 & 31.83 & 48.00 & 49.00 & 47.00 & 21.97 & 3.00 & 1.00 & 12.76 & 10.06 & 12.85  \\
    & Standard CSE   &  27.73 & 21.08 & 7.61 & 42.01 & 28.17 & 19.78 & 37.74 & 34.01 & 27.14 & 46.00 & 55.00 & 33.00 & 31.16 & 21.36 & 16.02 & 22.50 & 23.79 & 21.97  \\
    &  \quad $+$ Individual Cache &  29.93 & 27.66 & 14.93 & 55.10 & 40.75 & 45.48 & 45.88 & 46.10 & 32.07 & 50.00 & 64.00 & 57.00 & 61.99 & 61.69 & 65.64 & 27.99 & 30.44 & 34.19  \\
    &  \quad $+$ Shared Cache  &     30.93 & 27.14 & 19.18 & 54.34 & 40.53 & 45.96 & 45.71 & 45.37 & 35.53 & 50.00 & 59.00 & 52.00 & 56.19 & 48.16 & 35.69 & 26.61 & 29.15 & 31.30  \\
    & H2O $+$ Shared Cache &     29.41 & 23.47 & 18.85 & 53.28 & 38.04 & 42.23 & 45.99 & 45.70 & 34.21 & 48.00 & 62.00 & 52.00 & 57.83 & 43.32 & 42.67 & 7.17 & 4.87 & 4.72  \\


    \bottomrule
    \end{tabular}
    }

\caption{The detailed results on six different long sequence tasks, where $l_s=256, k=768$ for all methods. Results are separately presented by grouping text with different source lengths.}
    \label{tab:benchmark}
\end{table*}

\begin{table*}[t]
  \centering
  \resizebox{0.99\linewidth}{!}{
    \begin{tabular}{llrrrrrrrrrrrrrrrrrrrrrr}
    \toprule
      \multirow{2}*{Models} & \multirow{2}*{Settings} & \multicolumn{3}{c}{Qasper} & \multicolumn{3}{c}{MultifieldQA} & \multicolumn{3}{c}{HotpotQA} & \multicolumn{3}{c}{Trec}  &\multicolumn{3}{c}{TriviaQA} & \multicolumn{3}{c}{SamSum}   \\
      \cmidrule(r{4pt}){3-5} \cmidrule(r{4pt}){6-8} \cmidrule(r{4pt}){9-11}\cmidrule(r{4pt}){12-14}\cmidrule(r{4pt}){15-17} \cmidrule(r{4pt}){18-20}
       & & 0-4k & 4k-8k & 8k+ & 0-4k & 4-8k  & 8k+ & 0-4k & 4-8k & 8k+ & 0-4k & 4-8k & 8k+ & 0-4k & 4-8k & 8k+ & 0-4k &  4-8k & 8k+ \\
    \midrule
        \multirow{3}*{\makecell[l]{Llama 2 \\ 7B Chat}}
    & Standard CSE &  11.70 & 14.56 & 25.00 & 22.51 & 19.57 & 8.63 & 29.64 & 24.84 & 34.76 & 48.00 & 48.00 & 42.00 & 74.15 & 73.14 & 73.13 & 35.49 & 26.95 & 26.87 \\
    & \quad $+$ Individual Cache & 18.46 & 12.47 & 31.89 & 36.58 & 29.15 & 14.87 & 39.26 & 38.63 & 38.60 & 46.00 & 57.00 & 58.00 & 68.07 & 80.72 & 79.66 & 37.73 & 31.15 & 27.81 \\
    & \quad $+$ Shared Cache &  19.48 & 11.82 & 34.78 & 35.49 & 31.19 & 14.34 & 41.13 & 37.29 & 40.58 & 49.00 & 55.00 & 54.50 & 64.71 & 84.09 & 87.52 & 37.22 & 31.33 & 29.67 \\
    \midrule
    \multirow{3}*{\makecell[l]{Llama 2\\ 13B Chat}}
    & Standard CSE & 17.57 & 12.26 & 3.13 & 29.76 & 15.33 & 6.94 & 37.94 & 23.75 & 13.13 & 52.00 & 47.00 & 41.00 & 84.62 & 69.46 & 41.09 & 34.39 & 20.98 & 13.63  \\
    & \quad $+$ Individual Cache & 19.29 & 23.11 & 1.69 & 38.49 & 20.53 & 10.49 & 38.26 & 26.90 & 22.27 & 53.00 & 59.00 & 55.00 & 85.48 & 79.79 & 81.49 & 34.71 & 28.20 & 20.33 \\
    & \quad $+$ Shared Cache  & 20.36 & 21.80 & 5.90 & 38.89 & 21.18 & 12.75 & 40.20 & 27.66 & 22.26 & 54.00 & 61.00 & 49.00 & 85.39 & 78.56 & 69.72 & 33.15 & 24.04 & 14.97  \\
            \midrule
    \multirow{3}*{\makecell[l]{Mistral\\ 7B Instruct}}
    & Standard CSE & 27.48 & 21.83 & 9.61 & 38.67 & 28.22 & 17.08 & 37.20 & 28.43 & 25.93 & 42.00 & 44.00 & 29.00 & 32.82 & 26.86 & 28.51 & 25.69 & 20.80 & 21.36  \\
    &  \quad $+$ Individual Cache & 28.93 & 25.32 & 14.08 & 53.46 & 36.87 & 47.70 & 52.38 & 45.48 & 38.90 & 44.00 & 55.00 & 53.00 & 61.55 & 64.69 & 64.02 & 30.46 & 32.94 & 32.25 \\ 
    &  \quad $+$ Shared Cache  & 29.83 & 25.22 & 18.90 & 53.68 & 37.75 & 46.49 & 50.84 & 45.40 & 35.89 & 42.00 & 53.00 & 46.00 & 60.27 & 45.66 & 32.02 & 31.28 & 33.34 & 33.75   \\
    \bottomrule
    \end{tabular}
    }

\caption{The detailed results on six different long sequence tasks, where $l_s=512, k=512$ for all methods. Results are separately presented by grouping text with different source lengths.}
    \label{tab:benchmark_l=512_k=512}
\end{table*}

\begin{table*}[t]
  \centering
  \resizebox{0.99\linewidth}{!}{
    \begin{tabular}{llrrrrrrrrrrrrrrrrrrrrrr}
    \toprule
      \multirow{2}*{Models} & \multirow{2}*{Settings} & \multicolumn{3}{c}{Qasper} & \multicolumn{3}{c}{MultifieldQA} & \multicolumn{3}{c}{HotpotQA} & \multicolumn{3}{c}{Trec}  &\multicolumn{3}{c}{TriviaQA} & \multicolumn{3}{c}{SamSum}   \\
      \cmidrule(r{4pt}){3-5} \cmidrule(r{4pt}){6-8} \cmidrule(r{4pt}){9-11}\cmidrule(r{4pt}){12-14}\cmidrule(r{4pt}){15-17} \cmidrule(r{4pt}){18-20}
       & & 0-4k & 4k-8k & 8k+ & 0-4k & 4-8k  & 8k+ & 0-4k & 4-8k & 8k+ & 0-4k & 4-8k & 8k+ & 0-4k & 4-8k & 8k+ & 0-4k &  4-8k & 8k+ \\
    \midrule
        \multirow{3}*{\makecell[l]{Llama 2\\ 7B Chat}}
    & Standard CSE &  10.26 & 10.42 & 29.17 & 24.05 & 20.45 & 8.13 & 31.03 & 23.94 & 24.82 & 33.00 & 35.50 & 28.00 & 71.38 & 74.90 & 68.84 & 25.82 & 21.00 & 18.44 \\
    & \quad $+$ Individual Cache & 15.22 & 11.09 & 28.19 & 28.43 & 21.79 & 18.14 & 44.69 & 28.28 & 36.19 & 31.00 & 49.00 & 40.00 & 58.09 & 73.99 & 66.09 & 30.96 & 18.57 & 14.64 \\
    & \quad $+$ Shared Cache & 17.50 & 13.23 & 35.97 & 30.51 & 22.44 & 9.29 & 44.98 & 34.58 & 34.14 & 30.00 & 47.00 & 37.00 & 62.56 & 76.12 & 68.62 & 31.14 & 20.63 & 22.61 \\
    \midrule
    \multirow{3}*{\makecell[l]{Llama 2\\ 13B Chat}}
    & Standard CSE & 11.76 & 4.74 & 5.72 & 21.47 & 8.90 & 5.64 & 19.25 & 4.87 & 11.98 & 41.00 & 41.00 & 28.00 & 69.39 & 35.10 & 35.48 & 24.52 & 3.98 & 2.67 \\
    & \quad $+$ Individual Cache & 18.02 & 10.93 & 12.26 & 32.66 & 20.36 & 15.97 & 28.30 & 23.30 & 26.32 & 44.00 & 48.00 & 41.00 & 84.61 & 78.51 & 81.02 & 26.34 & 7.97 & 7.29 \\
    & \quad $+$ Shared Cache  & 17.27 & 10.20 & 10.85 & 37.06 & 22.92 & 17.46 & 26.75 & 26.69 & 26.45 & 42.00 & 49.00 & 37.00 & 84.08 & 78.79 & 76.02 & 26.76 & 8.72 & 6.33 \\
            \midrule
    \multirow{3}*{\makecell[l]{Mistral\\ 7B Instruct}}
    & Standard CSE & 25.59 & 20.65 & 18.20 & 32.87 & 26.97 & 19.38 & 34.74 & 22.01 & 25.67 & 27.00 & 25.00 & 19.00 & 38.67 & 24.65 & 25.08 & 24.72 & 19.83 & 21.33 \\
    &  \quad $+$ Individual Cache & 26.13 & 20.69 & 18.89 & 48.67 & 37.45 & 47.71 & 48.60 & 41.52 & 33.23 & 33.00 & 48.00 & 33.00 & 57.99 & 43.55 & 42.02 & 30.51 & 28.78 & 23.36 \\
    &  \quad $+$ Shared Cache  & 25.67 & 21.22 & 17.88 & 50.88 & 36.95 & 41.19 & 45.26 & 41.69 & 36.48 & 33.00 & 46.00 & 31.00 & 47.63 & 21.22 & 19.52 & 30.90 & 27.91 & 26.56 \\
    \bottomrule
    \end{tabular}
    }

\caption{The detailed results on six different long sequence tasks, where $l_s=768, k=256$ for all methods. Results are separately presented by grouping text with different source lengths.}
    \label{tab:benchmark_l=768_k=256}
\end{table*}

\begin{table*}[t]
  \centering
  \resizebox{0.99\linewidth}{!}{
    \begin{tabular}{llrrrrrrrrrrrrrrrrrrrrrr}
    \toprule
      \multirow{2}*{Models} & \multirow{2}*{Settings} & \multicolumn{3}{c}{Qasper} & \multicolumn{3}{c}{MultifieldQA} & \multicolumn{3}{c}{HotpotQA} & \multicolumn{3}{c}{Trec}  &\multicolumn{3}{c}{TriviaQA} & \multicolumn{3}{c}{SamSum}   \\
      \cmidrule(r{4pt}){3-5} \cmidrule(r{4pt}){6-8} \cmidrule(r{4pt}){9-11}\cmidrule(r{4pt}){12-14}\cmidrule(r{4pt}){15-17} \cmidrule(r{4pt}){18-20}
       & & 0-4k & 4k-8k & 8k+ & 0-4k & 4-8k  & 8k+ & 0-4k & 4-8k & 8k+ & 0-4k & 4-8k & 8k+ & 0-4k & 4-8k & 8k+ & 0-4k &  4-8k & 8k+ \\
    \midrule
        \multirow{3}*{\makecell[l]{Llama 2\\ 7B Chat}}
    & Standard CSE &  10.15 & 13.93 & 25.00 & 25.91 & 21.59 & 16.65 & 32.68 & 31.54 & 33.44 & 55.00 & 56.00 & 51.00 & 71.75 & 73.64 & 73.72 & 33.52 & 28.26 & 26.42 \\
    & \quad $+$ Individual Cache & 19.97 & 14.91 & 29.33 & 32.87 & 28.94 & 24.18 & 43.58 & 39.46 & 42.37 & 53.00 & 61.00 & 64.00 & 68.72 & 81.19 & 85.60 & 37.44 & 31.83 & 32.00 \\
    & \quad $+$ Shared Cache & 19.63 & 15.07 & 30.28 & 33.19 & 30.11 & 27.17 & 41.13 & 38.12 & 43.18 & 57.00 & 60.00 & 63.00 & 69.41 & 82.80 & 86.93 & 37.90 & 33.00 & 29.70  \\
    \midrule
    \multirow{3}*{\makecell[l]{Mistral\\ 7B Instruct}}
    & Standard CSE   & 22.96 & 20.93 & 20.94 & 41.62 & 27.26 & 18.66 & 33.26 & 26.06 & 19.30 & 46.00 & 55.00 & 45.00 & 33.71 & 22.68 & 24.65 & 14.41 & 15.45 & 17.76  \\
    &  \quad $+$ Individual Cache & 28.76 & 25.26 & 14.45 & 55.43 & 36.75 & 43.01 & 39.75 & 41.59 & 34.00 & 48.00 & 59.00 & 54.00 & 56.79 & 57.22 & 63.82 & 28.02 & 32.70 & 34.20 \\
    &  \quad $+$ Shared Cache  &  29.15 & 25.13 & 15.74 & 54.55 & 37.05 & 37.78 & 43.19 & 43.81 & 32.15 & 46.00 & 57.00 & 52.00 & 53.79 & 45.32 & 40.52 & 27.92 & 31.97 & 32.76 \\
    \bottomrule
    \end{tabular}
    }

\caption{The detailed results on six different long sequence tasks, where $l_s=512, k=768$ for all methods. Results are separately presented by grouping text with different source lengths..}
    \label{tab:benchmark_l=512_k=768}
\end{table*}

\begin{table*}[t]
  \centering
  \resizebox{0.99\linewidth}{!}{
    \begin{tabular}{llrrrrrrrrrrrrrrrrrrrrrr}
    \toprule
      \multirow{2}*{Models} & \multirow{2}*{Settings} & \multicolumn{3}{c}{Qasper} & \multicolumn{3}{c}{MultifieldQA} & \multicolumn{3}{c}{HotpotQA} & \multicolumn{3}{c}{Trec}  &\multicolumn{3}{c}{TriviaQA} & \multicolumn{3}{c}{SamSum}   \\
      \cmidrule(r{4pt}){3-5} \cmidrule(r{4pt}){6-8} \cmidrule(r{4pt}){9-11}\cmidrule(r{4pt}){12-14}\cmidrule(r{4pt}){15-17} \cmidrule(r{4pt}){18-20}
       & & 0-4k & 4k-8k & 8k+ & 0-4k & 4-8k  & 8k+ & 0-4k & 4-8k & 8k+ & 0-4k & 4-8k & 8k+ & 0-4k & 4-8k & 8k+ & 0-4k &  4-8k & 8k+ \\
    \midrule
        \multirow{3}*{\makecell[l]{Llama 2\\ 7B Chat}}
    & Standard CSE &  8.82 & 14.12 & 29.17 & 26.33 & 21.70 & 16.74 & 34.49 & 33.97 & 33.62 & 58.00 & 52.00 & 53.00 & 75.62 & 78.95 & 73.62 & 33.17 & 28.76 & 24.21 \\
    & \quad $+$ Individual Cache & 22.65 & 16.52 & 25.05 & 34.04 & 30.62 & 18.85 & 38.23 & 36.64 & 40.67 & 53.00 & 61.00 & 59.00 & 69.16 & 82.77 & 82.37 & 36.55 & 33.17 & 27.59 \\
    & \quad $+$ Shared Cache & 20.52 & 15.11 & 24.83 & 33.33 & 27.58 & 20.70 & 37.96 & 37.94 & 42.66 & 53.00 & 63.00 & 60.00 & 70.13 & 81.68 & 84.47 & 37.59 & 30.34 & 29.89 \\
    \midrule
    \multirow{3}*{\makecell[l]{Mistral\\ 7B Instruct}}
    & Standard CSE   & 26.24 & 22.10 & 17.73 & 42.75 & 26.37 & 16.64 & 35.25 & 26.72 & 19.11 & 48.00 & 51.00 & 38.00 & 31.18 & 15.87 & 16.21 & 16.66 & 12.35 & 15.37 \\
    &  \quad $+$ Individual Cache & 28.28 & 23.91 & 16.08 & 54.69 & 36.15 & 35.06 & 41.04 & 38.40 & 29.53 & 45.00 & 59.00 & 54.00 & 56.79 & 50.32 & 56.02 & 28.03 & 30.79 & 33.54 \\
    &  \quad $+$ Shared Cache & 28.99 & 23.95 & 15.14 & 55.93 & 37.74 & 35.83 & 42.36 & 39.44 & 29.61 & 47.00 & 56.00 & 52.00 & 53.79 & 44.82 & 39.69 & 28.46 & 30.34 & 29.11    \\
    \bottomrule
    \end{tabular}
    }

\caption{The detailed results on six different long sequence tasks, where $l_s=768, k=768$ for all methods. Results are separately presented by grouping text with different source lengths.}
    \label{tab:benchmark_l=768_k=768}
\end{table*}

\section{Statistics for Each Dataset}
\label{app:dataset_statistic}

\paragraph{Qasper} \citep{dasigi2021dataset} consists of 5049 questions from 1585 NLP research papers. The questions are created by practitioners who read only the title and abstract, and answered by another group, who also provide supporting evidence.
We use all available questions for each of the 224 documents selected by \citep{bai2023longbench} from this dataset to evaluate model performance. When doing the intersection probing experiments, we use all 416 documents from the test split of Qasper. We randomly choose one question as the instruction text for each document. 

\paragraph{MultifieldQA} \citep{bai2023longbench} consists of long articles from about 10 sources, including Latex papers, judicial documents, government work reports, and PDF documents indexed by Google. For each long article, several PhD and master students are invited to annotate. Each annotator is asked to propose questions with definitive answers as much as possible. We only use the English 
version of this dataset in our experiments. It contains 150 long documents.
\paragraph{HotpotQA} \citep{yang2018hotpotqa} is a dataset with 113,000 question-answer pairs based on Wikipedia. This dataset requires multi-document reasoning to answer questions, and the questions are quite diverse and not tied to specific knowledge bases. HotpotQA has been adapted by \citep{bai2023longbench} for long context evaluation, by concatenating the evidence text containing the answer along with several distracting articles. We use all 150 documents from the adapted HotpotQA for our experiments.

\paragraph{TriviaQA} \citep{joshi2017triviaqa} is a reading comprehension dataset containing over 650K question-answer-evidence triples. Averagely, six evidence documents are collected for each question. We use all 300 document-question pairs selected by \citep{bai2023longbench}, where each document consists of the concatenation of all available evidence documents for that question.

\paragraph{TREC} \citep{li-roth-2002-learning} is a question type classification dataset collected from 4500 English questions published by USC \citep{hovy-etal-2001-toward} together with 500 manually constructed questions for a few rare question types. This dataset has also been adapted for long context evaluation \citep{bai2023longbench}. This is achieved by sampling several cases from the training set to create few-shot examples as long context. We use all 300 examples from the adapted TREC.

\paragraph{SamSum}\citep{gliwa2019samsum}
includes around 16K messenger-like conversations with summaries, created by English-fluent linguists. These conversations mirror the topics and styles of real-life messenger interactions, ranging from informal to formal, and may include slang, emoticons, and typos. Each conversation is annotated with a third-person summary, providing a concise overview of the discussion. This dataset has been adapted for long context evaluation as well in the same manner as the TREC dataset, and we use all 300 examples from this adaptation.

\paragraph{PG19} \citep{rae2019compressive} includes a set of books extracted from the Project Gutenberg books library, that were published before 1919. We concatenate several selected books from this dataset to form a super long document and test the language modeling ability of our proposed methods on this document to up to 400K tokens in length.


\section{Detailed Results for Each Dataset}
\label{app:results_dataset}

In this section, we provide the dataset results for all the experiments. 
In Table~\ref{tab:benchmark_avg}, we show the averaged results of all the baseline models and the \modelname{} model, while the detailed results containing different datasets are shown in Table~\ref{tab:benchmark}. Dataset-wise experiment results using different hyperparameters are shown in Table~\ref{tab:benchmark_l=512_k=512}, Table~\ref{tab:benchmark_l=768_k=256}, Table~\ref{tab:benchmark_l=512_k=768}.  
Table~\ref{tab:benchmark_l=768_k=768},
Table~\ref{tab:benchmark_l=64_k=768},
and Table~\ref{tab:start_size_details}.

\section{Information Neglect of the Sliding Window Methods}
\label{app:sliding}
As pointed out by \citet{jiang2023mistral}, the sliding window method with a window size of $w$ would make the $i$\textit{th} token representation $h_i^l$ in a specific layer $l$ access tokens from the input layer at a distance of up to $l\times w$. 
This is due to the inherent design of attention mechanism, where the representations of a former token in one layer could only be aggregated to the representations of a following token in the next layer. We describe this phenomenon more specifically by analyzing the equation of the sliding window attention mechanism for the token $t_i$ with index $i$ in a specific layer $l$,
\begin{equation}
    a^l_{ij} = \frac{\exp\left(\frac{q^l_i \cdot k^l_j}{\sqrt{d_k}}\right)}{\sum_{j'=i-w}^i \exp\left(\frac{q^l_i \cdot k^l_{j'}}{\sqrt{d_k}}\right)}
\end{equation}
\begin{equation}
   \mathrm{Attention}^l_i = \sum_{j=i-w}^{i} a^l_{ij} \cdot v^l_j
\end{equation}
where $d_k$ is the dimension of the hidden states, $i$ and $j$ are the the indexes of the query token and the tokens whose information are aggregated, respectively. 
As all the tokens are processed parallelly in one layer, 
the hidden states $v_j$ and $k_j$ could only contain their aggregated information from the previous layer, acquired by $\mathrm{Attention}^{l-1}_j$. 
Considering $q_i$ could only attend to $\{v_{i - w}, \dots, v_i\}$ and $\{k_{i - w}, \dots, k_i\}$ in one layer, the information aggregation range $r(i,j,l,l')$ for $t_i$ from layer $l'$ to $l$ is,
\begin{equation}
r(i,j,l,l') = \bigcup_{l^*=l'}^l \left\{ \bigcup_{i^*=i - (l - l^*) \times w}^i  \mathrm{Attention}^{l^*}_{i^*} \right\}
\end{equation}

Hence, the information of token $t_i$ in the layer $0$~(i.e., the embedding layer) would completely disappear in layer $l$ after $l\times w$ time steps. Considering the effect that LLM would use specific layers to process the specifc information (e.g., syntax, task vector, etc)~\citep{hendel2023incontext, todd2023function}, the specific information for one token might disappear merely after a few window lengths.

\section{Supplemental Experiments for Information Neglect Problem}
\label{app:full_vs_stateeviction}
We discussed the issue of information neglect in Section~\ref{sec:information_neglect}. 
In this section, we present a straightforward experiment to further demonstrate the existence of this problem. 
Specifically, we compare the performance of models that read the full context of the document with those employing state eviction techniques. This experiment utilizes the Llama 3 8B Instruct model across the six reading comprehension datasets mentioned in our paper. As most long documents exceed the model's processing capacity, we limited our tests to examples with fewer than 4096 tokens. Additionally, we applied 8-bit quantization for efficiency. Alongside the previously discussed state eviction models, we also include our proposed CItruS model. We set $k=256,l_s=256$ for these experiments to simulate scenarios with small caches and long documents. The results are shown in Table~\ref{tab:supplement_information_neglect}:

\begin{table*}[t]
  \centering
  \resizebox{0.90\linewidth}{!}{
    \begin{tabular}{l|rr|rrrrrrrrrrrrrrrrrrrrr}
    \toprule
  Cache Type & Avg. Rank & Avg. Results & Qasper & Multifieldqa\_En & Hotpot QA & TREC & TriviaQA & SamSum \\
  \midrule

     Streaming LLM  & 2.83 & 33.94 & 23.58 & 16.25 & 38.27 & 44.29 & 46.39 & 34.85 \\
     TOVA  & 3.00 & 36.07 & 15.50 & 17.60 & 44.84 & 47.14 & 57.10 & 34.26  \\
     RoCo  & 2.50 & 34.00 & 21.83 & 25.47 & 28.33 & 42.86 & 51.93 & 33.58 \\
     H2O  &  4.33 & 37.96 & 16.39 & 25.69 & 40.20 & 54.29 & 62.17 & 29.03 \\
    \midrule
      Standard CSE &  3.33 & 35.84 & 15.87 & 18.48 & 39.02 & 44.29 & 60.25 & 37.12  \\
     \quad $+$ Individual Cache  &  7.00 & 49.12 & 23.17 & 46.43 & 50.95 & 47.14 & 88.89 & 38.12  \\
     \quad $+$ Shared Cache & 7.67 & 49.82 & 26.73 & 47.61 & 49.90 & 48.57 & 88.89 & 37.20 \\
     H2O $+$ Shared Cache & 5.83 & 45.97 & 18.63 & 39.44 & 48.06 & 48.57 & 85.44 & 35.66  \\
     \midrule
     Full Text & 7.67 & 49.87 & 23.97 & 52.29 & 46.56 & 54.29 & 83.45 & 38.65  \\
    \bottomrule
    \end{tabular}
    }

\caption{The average results and the average reversed ranks of the Llama 3 8B Instruct model on six different long sequence tasks, where Avg. Rank represents the averaged reversed rank and Avg. Results represents the averaged results. }
    \label{tab:supplement_information_neglect}
\end{table*}

Results show: 1) There is a large gap between the performance of the previous cache eviction methods and the model that could “read” the full text. 2) We would like to point out that this is not the ideal case for our proposed CItruS, which is designed for processing long sequences beyond the capacity of LLMs. However, even with the short context, the proposed method approaches the performance of full-context models better than the baseline models. Notably, in the TriviaQA and Qasper datasets, CItruS outperforms the models with the full text. We hypothesize that it is because some noisy information is eliminated during the eviction process.

\section{Prompt for Each Task}
\label{app:prompt}
We show all of the prompts we used for each task in Table~\ref{tab:template}.

\begin{table}[t]
  \centering

  \resizebox{0.99\linewidth}{!}{
    \begin{tabular}{lp{10cm}}
    \toprule
    Datasets & Prompt \\
    \midrule     
   
    Qasper & \textcolor{blue}{You are given a scientific article and a question. Answer the question as concisely as you can, using a single phrase or sentence if possible. If the question cannot be answered based on the information in the article, write ``unanswerable''. If the question is a yes/no question, answer ``yes'', ``no'', or ``unanswerable''. Do not provide any explanation.\textbackslash n\textbackslash nArticle: \{context\}\textbackslash n\textbackslash n}\textcolor{red}{Answer the question based on the above article as concisely as you can, using a single phrase or sentence if possible. If the question cannot be answered based on the information in the article, write ``unanswerable''. If the question is a yes/no question, answer ``yes'', ``no'', or ``unanswerable''. Do not provide any explanation.\textbackslash n\textbackslash n Question: \{input\}\textbackslash n\textbackslash n Answer:}\\    \midrule 
    MultifieldQA &  \textcolor{blue}{Read the following text and answer briefly.\textbackslash n\textbackslash n\{context\}\textbackslash n\textbackslash n}\textcolor{red}{Now, answer the following question based on the above text, only give me the answer and do not output any other words.\textbackslash n\textbackslash nQuestion: \{input\}\textbackslash n Answer:} \\    \midrule 
    HotpotQA & \textcolor{blue}{Answer the question based on the given passages. Only give me the answer and do not output any other words.\textbackslash n\textbackslash nThe following are given passages.\textbackslash n\textbackslash n\{context\}\textbackslash n\textbackslash n } \textcolor{red}{Answer the question based on the given passages. Only give me the answer and do not output any other words.\textbackslash n\textbackslash nQuestion: \{input\}\textbackslash n Answer:} \\    \midrule 
    
    TriviaQA & \textcolor{blue}{Answer the question based on the given passage. Only give me the answer and do not output any other words. The following are some examples.\textbackslash n\textbackslash n\textbackslash n\textbackslash n\{context\}\textbackslash n\textbackslash n\textbackslash n\textbackslash n} \textcolor{red}{Question: \{input\}\textbackslash n\textbackslash n\textbackslash n\textbackslash nAnswer:} \\    \midrule 
     TREC & \textcolor{blue}{Please determine the type of the question below. Here are some examples of questions.\textbackslash n\textbackslash n\textbackslash n\textbackslash n\{context\}\textbackslash n\textbackslash n}\textcolor{red}{\{input\}} \\    \midrule 

    SamSum& \textcolor{blue}{Summarize the dialogue into a few short sentences. The following are some examples.\textbackslash n\textbackslash n\textbackslash n\textbackslash n\{context\}\textbackslash n\textbackslash n\textbackslash n\textbackslash n}\textcolor{red}{\{input\}} \\    \midrule 

    Passkey Retrieval& \textcolor{blue}{There is an important info hidden inside a lot of irrelevant text. Find it and memorize them. I will quiz you about the important information there.\{context\}\textbackslash n\textbackslash n\textbackslash n\textbackslash n}\textcolor{red}{What is the pass key? The pass key is } \\    \midrule 
    needle-in-a-haystack& \textcolor{blue}{system: You are a helpful AI bot that answers questions for a user. Keep your response short and direct \textbackslash n\textbackslash n user: \{context\}\textbackslash n\textbackslash n}\textcolor{red}{user: \{Question\} Don't give information outside the document or repeat your findings\textbackslash n\textbackslash n system:} \\
    \bottomrule
    \end{tabular}
    }
  \caption{The prompt used in our experiments. Text in \textcolor{blue}{blue} represents the context while text in \textcolor{red}{red} represents the instruction we used.}
    \label{tab:template}
\end{table}

\begin{table}[t]
  \centering

  \resizebox{0.99\linewidth}{!}{
    \begin{tabular}{lp{10cm}}
    \toprule
    Datasets & Prompt \\
    \midrule     
   
    Instruction 1 & Answer the question based on the given passages. Only give me the answer and do not output any other words.\textbackslash n\textbackslash nQuestion: How is the ground truth for fake news established?\textbackslash nAnswer:\\    \midrule 
    Instruction 2 & Answer the question based on the given passages. Only give me the answer and do not output any other words.\textbackslash n\textbackslash nQuestion: What architecture does the encoder have?\textbackslash nAnswer:\\    \midrule 
    Instruction 3 & Answer the question based on the given passages. Only give me the answer and do not output any other words.\textbackslash n\textbackslash nQuestion: Which case was brought to court first Miller v. California or Gates v. Collier ?\textbackslash nAnswer:\\    \midrule 
    Instruction 4 & Answer the question based on the given passages. Only give me the answer and do not output any other words.\textbackslash n\textbackslash nQuestion: What occupation is shared by both Marge Piercy and Richard Aldington?\textbackslash nAnswer:\\    \midrule 
    Instruction 5 & Answer the question based on the given passages. Only give me the answer and do not output any other words.\textbackslash n\textbackslash nQuestion: What is their definition of tweets going viral?\textbackslash nAnswer:\\    \midrule 
    Instruction 6 & Answer the question based on the given passages. Only give me the answer and do not output any other words.\textbackslash n\textbackslash nQuestion: Were any of these tasks evaluated in any previous work?\textbackslash nAnswer:\\    \midrule 
    Instruction 7 & Answer the question based on the given passages. Only give me the answer and do not output any other words.\textbackslash n\textbackslash nQuestion: What sentiment classification dataset is used?\textbackslash nAnswer:\\    \midrule 
    Instruction 8 & Answer the question based on the given passages. Only give me the answer and do not output any other words.\textbackslash n\textbackslash nQuestion: The historical Nimavar school in the Nimavar Bazaar, or bazar, is located in which country?\textbackslash nAnswer:\\    \midrule 
    Instruction 9 & Answer the question based on the given passages. Only give me the answer and do not output any other words.\textbackslash n\textbackslash nQuestion: For what type of work is the production company for The Year Without a Santa Claus best known?\textbackslash nAnswer:\\    \midrule 
    Instruction 10 & Answer the question based on the given passages. Only give me the answer and do not output any other words.\textbackslash n\textbackslash nQuestion: The physicist who is responsible for identifying the Rabi cycle won what award?\textbackslash nAnswer:\\
    \bottomrule
    \end{tabular}
    }
  \caption{The instruction used in the perplexity experiments. }
    \label{tab:ppl_instruction}
\end{table}
\section{Setup for the Needle-in-a-Haystack Task}
\label{app:needle}
Due to the computational cost limitation, we used one fact to conduct this task. The fact is ``The best thing to do in San Francisco is eat a sandwich and sit in Dolores Park on a sunny day.'' and the question input is ``What is the best thing to do in San Francisco?''. The document is concatenated from documents from Paul Graham Essays.
We cut the first 7 tokens where the model always generate ``The best thing to do in San Francisco is'' to avoid the miscalculation of the information overlap. The template we used is shown in Table~\ref{tab:template}.

\begin{table}[t]
  \centering
  \resizebox{0.8\linewidth}{!}{
    \begin{tabular}{lrrrrrrrrrrrrrrrrrrrrrr}
    \toprule
        Cache Type & 0-4k & 4-8k & 8k+  \\ \midrule
        Streaming LLM & 2.17 & 2.33 & 1.83  \\ 
        TOVA & 2.00 & 2.33 & 3.83  \\ 
        Roco & 3.67 & 4.67 & 5.33  \\ 
        H2O & 3.83 & 2.67 & 3.00  \\
        H2O $+$ Shared Cache & 6.67 & 6.33 & 4.83 \\
        Standard CSE & 3.33 & 3.00 & 3.33  \\ 
        \quad $+$ Individual Cache & 6.67 & 7.00 & \textbf{7.17} \\ 
        \quad $+$ Shared Cache & \textbf{7.17} & \textbf{7.33} & 6.33  \\ 
    \bottomrule
    \end{tabular}
    }

\caption{The averaged reversed rank results among all the 8 models on six different reading comprehension tasks with Llama 3 8B Instruct, where 8 is the highest score and 1 is the lowest score. Results are presented by grouping text with lengths of 0-4k, 4k-8k, and 8k$+$. Best results are bolded.}
    \label{tab:benchmark_llama3_rank}
\end{table}

\begin{table}[t]
  \centering
  \resizebox{0.8\linewidth}{!}{
    \begin{tabular}{lrrrrrrrrrrrrrrrrrrrrrr}
    \toprule
        Cache Type & 0-4k & 4-8k & 8k+  \\ \midrule
        Streaming LLM & 39.75 & 38.82 & 34.20  \\ 
        TOVA & 41.09 & 39.88 & 38.65  \\ 
        Roco & 45.17 & 45.10 & 42.41  \\ 
        H2O & 45.45 & 42.24 & 39.13  \\ 
        H2O  $+$ Shared Cache & 50.85 & 51.86 & 48.24 \\ 
        Standard CSE & 44.82 & 41.48 & 37.17  \\ 
        \quad $+$ Individual Cache & 50.66 & 52.10 & \textbf{53.03}  \\ 
        \quad $+$ Shared Cache & \textbf{51.17} & \textbf{53.20} & 51.66  \\ 
    \bottomrule
    \end{tabular}
    }

\caption{The averaged results among all the 8 models on six different reading comprehension tasks with Llama 3 8B Instruct, where 8 is the highest score and 1 is the lowest score. Results are presented by grouping text with lengths of 0-4k, 4k-8k, and 8k$+$. Best results are bolded.}
    \label{tab:benchmark_llama3_avg}
\end{table}

\section{Experimental Results with Llama 3}
\label{app:results_llama3}
We test the six reading comprehension tasks used in our paper with the newly released Llama 3 8B Instruct model. The results shown in Table~\ref{tab:benchmark_llama3_rank} and Table~\ref{tab:benchmark_llama3_avg} demonstrates that our proposed model continues to perform consistently better than other state eviction models.

\section{Experimental Results with BABILong Dataset}
\label{app:results_babilong}
We conduct supplementary experiments with BABILong~\citep{kuratov2024babilong}, a newly proposed dataset which contains long sequence needle-in-the-haystack tasks involving multiple supporting facts and requires the model to generate answers using multi-hop reasoning and temporal dynamics. 

We test our models and the baselines on the qa1, qa2, and qa3 subsets of BABILong with a maximum length of 128k tokens. All results were obtained using the Llama 3 8B Instruct model. The results are shown in Table~\ref{tab:babilong_qa1}, Table~\ref{tab:babilong_qa2}, and Table~\ref{tab:babilong_qa3}, where the “0k”, “1k”, and “64k” represent the context length of the subset.

Results show that our method performs better on these tasks, especially when the context length is longer. However, we want to point out that it is not guaranteed that our method could enhance the reasoning ability of LLMs. We are only claiming that our method can better help the state eviction methods retain more relevant information for downstream tasks when processing long sequences. The reasoning abilities depend on the model and how it leverages the information in the retained hidden states, which is fundamentally influenced by the pretraining process.
\begin{table*}[t]
  \centering
  \resizebox{0.99\linewidth}{!}{
    \begin{tabular}{llrrrrrrrrrrrrrrrrrrrrrr}
    \toprule
        Model & qa1\_0k & qa1\_1k & qa1\_2k & qa1\_4k & qa1\_8k & qa1\_16k & qa1\_32k & qa1\_64k & qa1\_128k & Avg\_qa1  \\ \midrule
        Streaming LLM & 0.98 & 0.83 & 0.74 & 0.43 & 0.25 & 0.16 & 0.08 & 0.03 & 0.03 & 0.39  \\ 
        TOVA & 0.98 & 0.83 & 0.68 & 0.48 & 0.41 & 0.29 & 0.15 & 0.04 & 0.02 & 0.43  \\ 
        Roco & 0.98 & 0.83 & 0.60 & 0.51 & 0.31 & 0.13 & 0.04 & 0.05 & 0.01 & 0.38  \\ 
        H2O & 0.98 & 0.83 & 0.31 & 0.20 & 0.13 & 0.05 & 0.01 & 0.01 & 0.01 & 0.28  \\ 
        H2O $+$ Shared Cache & 0.98 & 0.90 & 0.87 & 0.75 & 0.62 & 0.51 & 0.28 & 0.16 & 0.10 & 0.57  \\ 
        Standard CSE & 0.98 & 0.90 & 0.69 & 0.53 & 0.39 & 0.30 & 0.21 & 0.09 & 0.03 & 0.46  \\ 
        \quad $+$  Individual Cache & 0.98 & 0.89 & 0.82 & 0.73 & 0.63 & 0.56 & 0.50 & 0.31 & 0.26 & 0.63  \\ 
        \quad $+$  Shared Cache & 0.98 & 0.89 & 0.84 & 0.79 & 0.75 & 0.69 & 0.59 & 0.66 & 0.43 & \textbf{0.74} \\
    \bottomrule
    \end{tabular}
    }

\caption{The results on BABILong qa1 subset. Results are evaluated on test examples with different lengths. Best results are bolded.}
    \label{tab:babilong_qa1}
\end{table*}
\begin{table*}[t]
  \centering
  \resizebox{0.99\linewidth}{!}{
    \begin{tabular}{llrrrrrrrrrrrrrrrrrrrrrr}
    \toprule
        Model & qa2\_0k & qa2\_1k & qa2\_2k & qa2\_4k & qa2\_8k & qa2\_16k & qa2\_32k & qa2\_64k & qa2\_128k & Avg\_qa2  \\ \midrule
        Streaming LLM & 0.14 & 0.14 & 0.39 & 0.23 & 0.08 & 0.01 & 0.01 & 0.00 & 0.00 & 0.11  \\ 
        TOVA & 0.14 & 0.12 & 0.21 & 0.27 & 0.18 & 0.09 & 0.07 & 0.04 & 0.02 & 0.13  \\ 
        Roco & 0.14 & 0.10 & 0.02 & 0.29 & 0.11 & 0.05 & 0.03 & 0.01 & 0.01 & 0.08  \\ 
        H2O & 0.14 & 0.10 & 0.00 & 0.02 & 0.00 & 0.02 & 0.00 & 0.01 & 0.00 & 0.03  \\ \hline
        H2O $+$ Shared Cache & 0.14 & 0.12 & 0.11 & 0.05 & 0.01 & 0.00 & 0.00 & 0.00 & 0.00 & 0.05  \\ 
        Standard CSE & 0.14 & 0.13 & 0.12 & 0.27 & 0.23 & 0.10 & 0.05 & 0.04 & 0.02 & 0.12  \\ 
        \quad $+$ Individual Cache & 0.14 & 0.12 & 0.13 & 0.15 & 0.26 & 0.24 & 0.28 & 0.26 & 0.24 & \textbf{0.20}  \\
        \quad $+$  Shared Cache & 0.14 & 0.12 & 0.19 & 0.12 & 0.11 & 0.04 & 0.12 & 0.14 & 0.25 & 0.14 \\ 
    \bottomrule
    \end{tabular}
    }

\caption{The results on BABILong qa2 subset. Results are evaluated on test examples with different lengths. Best results are bolded.}
    \label{tab:babilong_qa2}
\end{table*}
\begin{table*}[t]
  \centering
  \resizebox{0.99\linewidth}{!}{
    \begin{tabular}{llrrrrrrrrrrrrrrrrrrrrrr}
    \toprule
        Model & qa3\_0k & qa3\_1k & qa3\_2k & qa3\_4k & qa3\_8k & qa3\_16k & qa3\_32k & qa3\_64k & qa3\_128k & Avg\_qa3  \\ \midrule
        Streaming LLM & 0.25 & 0.26 & 0.27 & 0.32 & 0.31 & 0.10 & 0.07 & 0.02 & 0.00 & 0.18  \\ 
        TOVA & 0.25 & 0.24 & 0.21 & 0.40 & 0.31 & 0.19 & 0.10 & 0.06 & 0.01 & 0.20  \\ 
        Roco & 0.25 & 0.26 & 0.02 & 0.19 & 0.23 & 0.12 & 0.07 & 0.04 & 0.01 & 0.13  \\ 
        H2O & 0.24 & 0.23 & 0.01 & 0.01 & 0.00 & 0.00 & 0.00 & 0.01 & 0.00 & 0.06  \\ 
        H2O $+$ Shared Cache & 0.24 & 0.23 & 0.22 & 0.07 & 0.04 & 0.02 & 0.02 & 0.01 & 0.01 & 0.10  \\ 
        Standard CSE & 0.25 & 0.23 & 0.19 & 0.32 & 0.28 & 0.16 & 0.11 & 0.07 & 0.04 & 0.18  \\ 
        \quad $+$  Individual Cache & 0.24 & 0.22 & 0.25 & 0.20 & 0.25 & 0.25 & 0.25 & 0.25 & 0.21 & 0.24  \\ 
        \quad $+$  Shared Cache & 0.24 & 0.22 & 0.26 & 0.19 & 0.19 & 0.22 & 0.32 & 0.33 & 0.34 & \textbf{0.26} \\
    \bottomrule
    \end{tabular}
    }

\caption{The results on BABILong qa3 subset. Results are evaluated on test examples with different lengths. Best results are bolded.}
    \label{tab:babilong_qa3}
\end{table*}
\section{Results of Perplexity with Other Instructions}
\label{app:other_instruction_ppl}
\begin{figure*}[b]
\centering
\includegraphics[width=1.99\columnwidth]{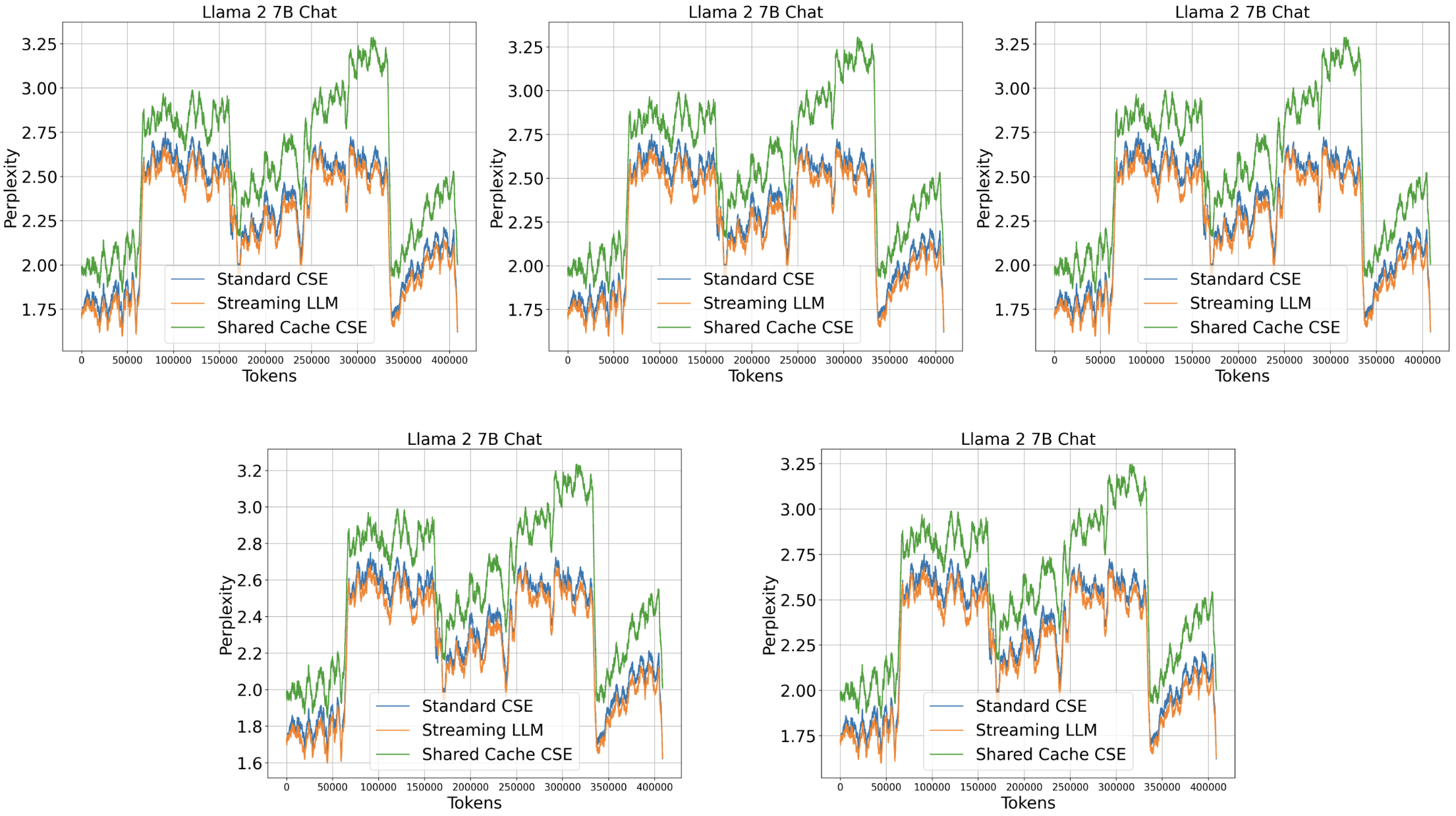}

\caption{The language modeling results on the Llama 2 7B chat model.  The instructions 1 to 5 listed in table \ref{tab:ppl_instruction} are used for the Shared Cache CSE method, respectively. The line chart is smoothed with a window size of 4096 for better visibility.}
\label{fig:ppl_llama_full1}
\end{figure*}

\begin{figure*}[t]
\centering

\includegraphics[width=1.99\columnwidth]{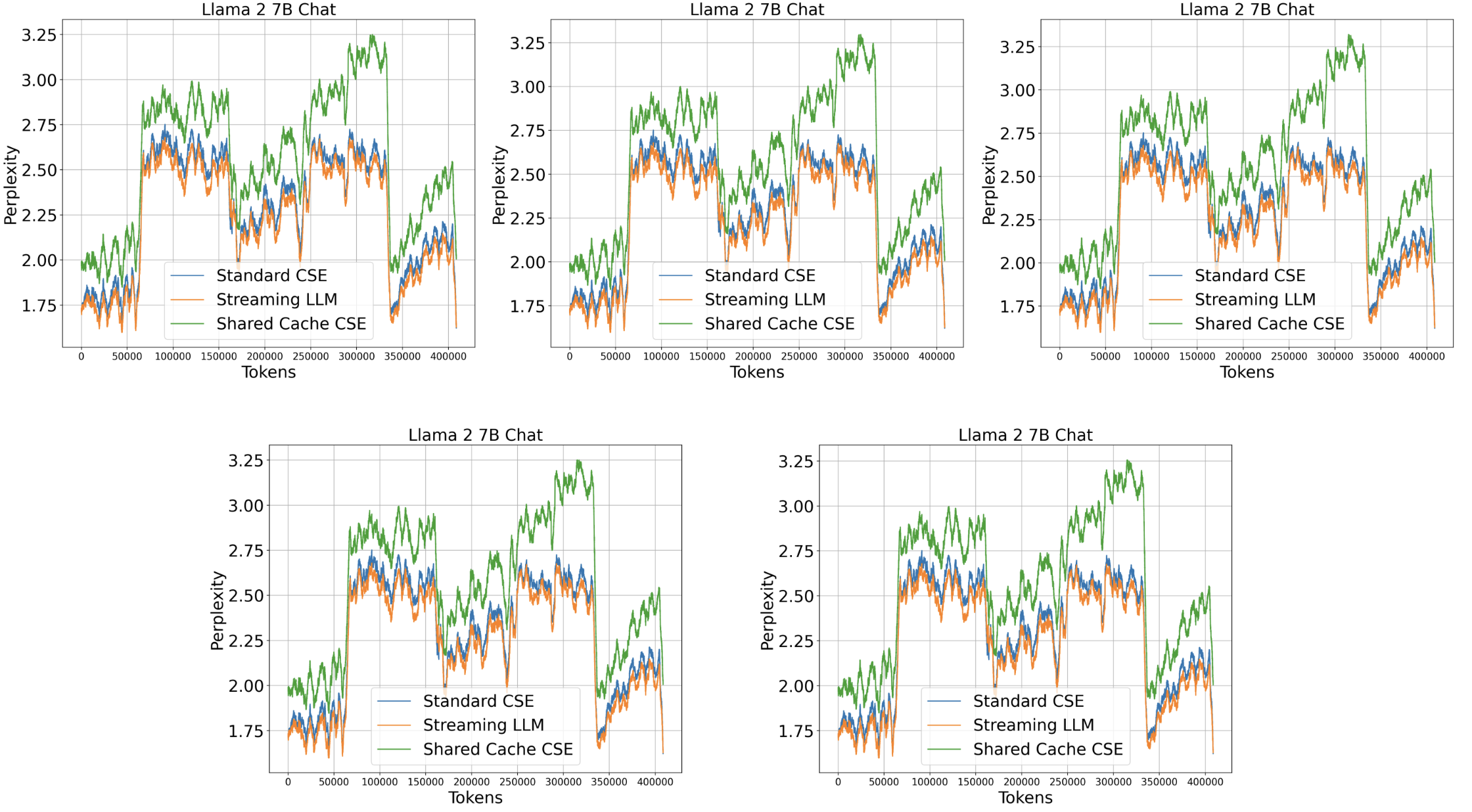}
\caption{The language modeling results on the Llama 2 7B chat model.  The instructions 6 to 10 listed in table \ref{tab:ppl_instruction} are used for the Shared Cache CSE method, respectively. The line chart is smoothed with a window size of 4096 for better visibility.}
\label{fig:ppl_llama_full2}
\end{figure*}

\begin{figure*}[t]
\centering

\includegraphics[width=1.99\columnwidth]{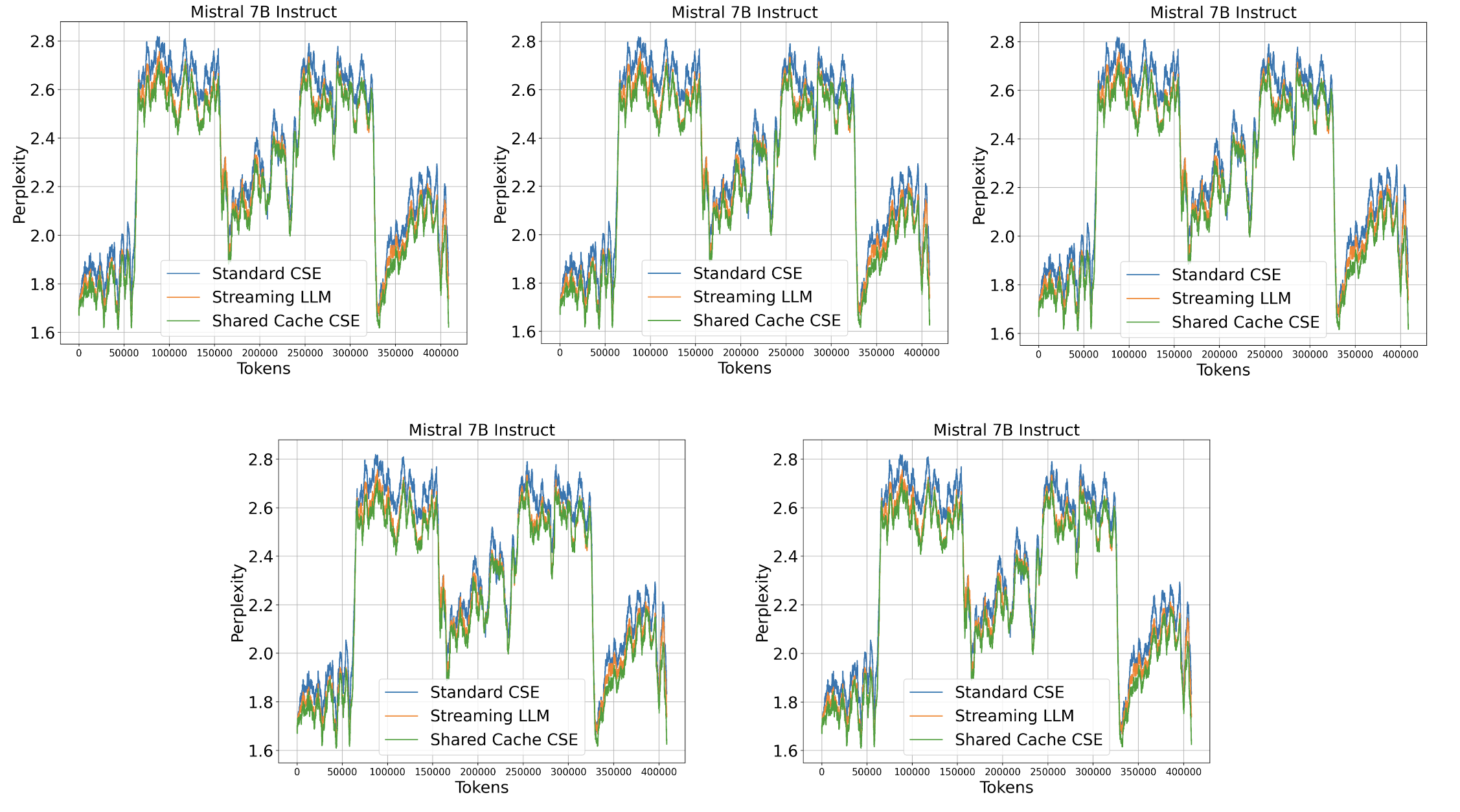}

\caption{The language modeling results on the Mistral 7B Instruct model. The instructions 1 to 5 listed in table \ref{tab:ppl_instruction} are used for the Shared Cache CSE method, respectively. The line chart is smoothed with a window size of 4096 for better visibility.}
\label{fig:ppl_mistral_full1}
\end{figure*}

\begin{figure*}[t]
\centering

\includegraphics[width=1.99\columnwidth]{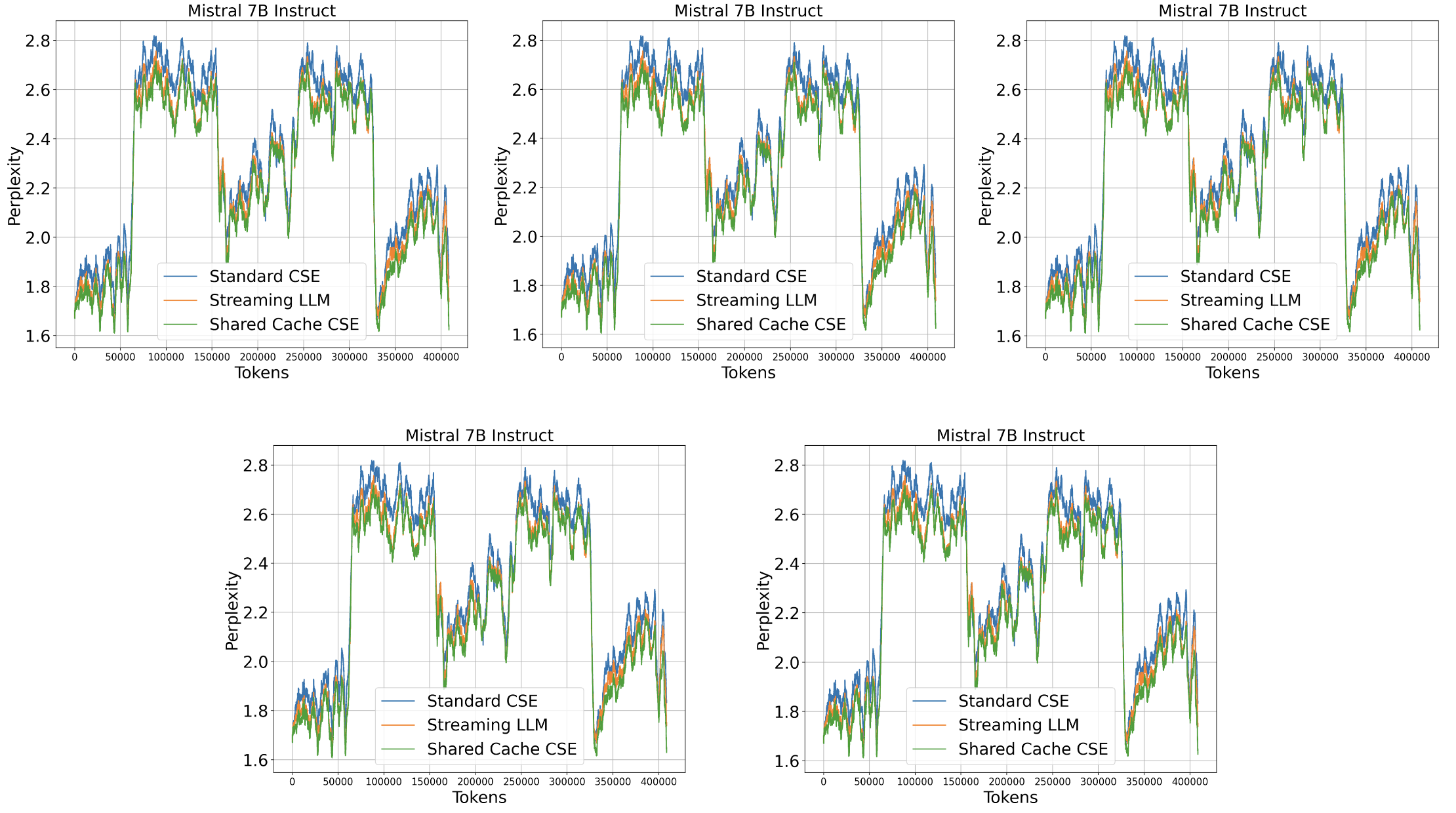}
\caption{The language modeling results on the Mistral 7B Instruct model. The instructions 6 to 10 listed in table \ref{tab:ppl_instruction} are used for the Shared Cache CSE method, respectively. The line chart is smoothed with a window size of 4096 for better visibility.}
\label{fig:ppl_mistral_full2}
\end{figure*}

We used 10 different instructions, shown in Table~\ref{tab:ppl_instruction}. We show the perplexity of models of \modelname{} with Shared cache when using these ten different instructions in Figure~\ref{fig:ppl_llama_full1}, Figure~\ref{fig:ppl_llama_full2}, Figure~\ref{fig:ppl_mistral_full1}, and Figure~\ref{fig:ppl_mistral_full2}. As these results demonstrate, the perplexity of our Shared Cache CSE remains consistent across a wide variety of instructions, similar to the standard CSE and streaming LLM methods.

\section{Discussion}
\label{app:discussion}
In this paper, we argue that the cache used in standard chunked state eviction (CSE) is primarily responsible for maintaining the perplexity of language models, whereas an instruction-aware cache offers advantages for long-sequence downstream tasks. This claim is supported by the following observations from our experiments: (1) perplexity evaluations and previous work on state eviction methods~\citep{zhang2024h2o, oren2024transformers} indicate that the basic cache effectively maintains language model perplexity; (2) performance improvements are observed when using an instruction-aware cache, which is only information that the model could access when generating the response during the task-solving thread. It is important to note that it is not solely the case that the standard cache only impacts perplexity while the instruction-aware cache solely affects task performance; there is potential overlapping, as demonstrated in our intersection calculation experiments discussed in Section \ref{sec:information_neglect}. However, the primary focus of these two types of caches remains distinct.

\section{Analysis on initial tokens}
\label{app:start_size}


\citet{xiao2023efficient} show that the initial tokens play a critical role in long-sequence language modeling by serving as ``attention sinks''. Although our proposed method does not specifically process the initial tokens, we assert that it can adaptively retain the hidden states of these tokens because they consistently receive a large proportion of attention weights. In this section, we conduct experiments that always preserve the first 4 initial tokens during the eviction process. 

Shown in Table~\ref{tab:start_size} and Table \ref{tab:start_size_details}, we demonstrate that the difference between our methods with and without the initial tokens are limited, showing the capability of keeping the ``attention sink'' tokens using our method.


\begin{table}[t]
  \centering
  \resizebox{0.99\linewidth}{!}{
    \begin{tabular}{lllrrrrrrrrr}
    \toprule
     \multirow{2}*{Param.} & \multirow{2}*{Settings}  & \multicolumn{3}{c}{Llama 2 7B} & \multicolumn{3}{c}{Mistral 7B} \\ 
    \cmidrule(r{4pt}){3-5}\cmidrule(r{4pt}){6-8}
    ~ & ~ & 0-4k & 4-8k & 8k+ & 0-4k & 4-8k & 8k+ \\
    \midrule
    \multirow{3}*{\makecell[l]{Start size $= 0$}}&     Standard CSE & 36.72 & 37.07 & 38.36 & 34.52 & 30.57 & 20.92  \\
    & \quad $+$ Individual Cache  &  \textbf{43.45} & 43.26 & 45.93 & \textbf{45.15} &\textbf{45.11} &\textbf{41.55} \\
    & \quad $+$ Shared Cache & 43.22 & \textbf{44.07} & \textbf{46.37} & 43.96 & 41.56 & 36.61 \\
    \midrule
    \multirow{3}*{\makecell[l]{Start size $= 4$}}& Standard CSE &  36.30 & 34.80 & 37.42 & 31.44 & 28.51 & 21.10  \\
    & \quad $+$ Individual Cache  &    \textbf{43.48} &\textbf{43.89}& 46.36 &\textbf{45.69} & \textbf{44.55} & \textbf{42.22}\\
    & \quad $+$ Shared Cache &  43.44 & 43.65 & \textbf{46.97}&  44.22 & 41.74 & 36.05  \\
    \bottomrule
    \end{tabular}
    }

\caption{Results of the different start sizes averaged on six different long sequence tasks. Best results are bolded. ``Param.'' stands for hyperparameters. }
    \label{tab:start_size}
\end{table}

\begin{table*}
  \centering
  \resizebox{0.99\linewidth}{!}{
    \begin{tabular}{llrrrrrrrrrrrrrrrrrrrrrr}
    \toprule
      \multirow{2}*{Models} & \multirow{2}*{Settings} & \multicolumn{3}{c}{Qasper} & \multicolumn{3}{c}{MultifieldQA} & \multicolumn{3}{c}{HotpotQA} & \multicolumn{3}{c}{Trec}  &\multicolumn{3}{c}{TriviaQA} & \multicolumn{3}{c}{SamSum}   \\
      \cmidrule(r{4pt}){3-5} \cmidrule(r{4pt}){6-8} \cmidrule(r{4pt}){9-11}\cmidrule(r{4pt}){12-14}\cmidrule(r{4pt}){15-17} \cmidrule(r{4pt}){18-20}
       & & 0-4k & 4k-8k & 8k+ & 0-4k & 4-8k  & 8k+ & 0-4k & 4-8k & 8k+ & 0-4k & 4-8k & 8k+ & 0-4k & 4-8k & 8k+ & 0-4k &  4-8k & 8k+ \\
    \midrule
        \multirow{3}*{\makecell[l]{Llama 2\\ 7B Chat}}
    & Standard CSE & 10.30 & 12.43 & 25.00 & 23.86 & 21.75 & 14.50 & 34.14 & 28.58 & 34.09 & 47.00 & 50.00 & 42.00 & 68.88 & 71.73 & 79.50 & 36.00 & 30.49 & 33.61 \\
    & \quad $+$ Individual Cache &  20.43 & 18.82 & 29.64 & 39.99 & 32.17 & 37.79 & 43.87 & 36.28 & 43.36 & 57.00 & 64.00 & 63.00 & 63.32 & 77.14 & 77.19 & 38.06 & 34.72 & 33.89 \\
    & \quad $+$ Shared Cache & 20.14 & 18.07 & 27.49 & 40.00 & 33.69 & 37.06 & 43.47 & 41.77 & 44.26 & 56.00 & 63.00 & 55.00 & 61.42 & 66.59 & 58.91 & 37.50 & 33.81 & 36.72 \\
    \midrule
    \multirow{3}*{\makecell[l]{Mistral\\ 7B Instruct}}
    & Standard CSE & 24.95 & 19.05 & 8.49 & 36.36 & 27.86 & 17.88 & 34.95 & 27.88 & 23.47 & 47.00 & 46.00 & 28.50 & 33.18 & 34.75 & 34.96 & 17.75 & 15.24 & 11.02 \\
    &  \quad $+$ Individual Cache &  32.00 & 28.45 & 15.80 & 57.42 & 39.76 & 47.02 & 47.02 & 41.63 & 30.00 & 51.00 & 64.00 & 56.00 & 59.43 & 61.32 & 67.01 & 28.40 & 28.90 & 25.38 \\
    &  \quad $+$ Shared Cache & 32.08 & 24.10 & 19.48 & 56.78 & 39.69 & 47.58 & 45.67 & 50.20 & 38.66 & 51.00 & 61.00 & 58.00 & 54.19 & 31.49 & 24.52 & 23.12 & 26.83 & 17.90 \\
    \bottomrule
    \end{tabular}
}

\caption{The detailed results on six different long sequence tasks, where $l_s=64, k=768$ for all methods. Results are separately presented by grouping text with different source lengths.}
\label{tab:benchmark_l=64_k=768}
\end{table*}

\begin{table*}[t]
  \centering
  \resizebox{0.99\linewidth}{!}{
    \begin{tabular}{llrrrrrrrrrrrrrrrrrrrrrr}
    \toprule
      \multirow{2}*{Models} & \multirow{2}*{Settings} & \multicolumn{3}{c}{Qasper} & \multicolumn{3}{c}{MultifieldQA} & \multicolumn{3}{c}{HotpotQA} & \multicolumn{3}{c}{Trec}  &\multicolumn{3}{c}{TriviaQA} & \multicolumn{3}{c}{SamSum}   \\
      \cmidrule(r{4pt}){3-5} \cmidrule(r{4pt}){6-8} \cmidrule(r{4pt}){9-11}\cmidrule(r{4pt}){12-14}\cmidrule(r{4pt}){15-17} \cmidrule(r{4pt}){18-20}
       & & 0-4k & 4k-8k & 8k+ & 0-4k & 4-8k  & 8k+ & 0-4k & 4-8k & 8k+ & 0-4k & 4-8k & 8k+ & 0-4k & 4-8k & 8k+ & 0-4k &  4-8k & 8k+ \\
    \midrule
        \multirow{3}*{\makecell[l]{Llama 2\\ 7B Chat}}
    & Standard CSE & 9.27 & 11.22 & 25.00 & 26.15 & 23.38 & 15.26 & 31.11 & 31.17 & 33.44 & 49.00 & 53.00 & 54.00 & 67.46 & 61.89 & 65.32 & 34.83 & 28.11 & 31.52 \\
    & \quad $+$ Individual Cache & 21.32 & 17.49 & 28.47 & 37.21 & 29.16 & 28.08 & 42.19 & 39.02 & 40.40 & 54.00 & 65.00 & 65.00 & 69.07 & 80.10 & 81.60 & 37.06 & 32.56 & 34.61 \\
    & \quad $+$ Shared Cache & 21.33 & 15.70 & 33.19 & 38.19 & 30.76 & 25.46 & 44.36 & 37.93 & 42.21 & 56.00 & 65.00 & 64.00 & 64.14 & 80.05 & 82.28 & 36.60 & 32.43 & 34.66 \\
    \midrule
    \multirow{3}*{\makecell[l]{Mistral\\ 7B Instruct}}
    & Standard CSE & 25.91 & 21.99 & 9.25 & 41.60 & 28.47 & 19.17 & 32.04 & 28.17 & 21.81 & 44.00 & 47.00 & 34.00 & 29.55 & 30.22 & 27.04 & 15.52 & 15.22 & 15.35 \\
    &  \quad $+$ Individual Cache & 30.37 & 27.09 & 15.30 & 56.32 & 40.28 & 47.21 & 47.74 & 45.98 & 33.80 & 49.00 & 64.00 & 56.00 & 61.99 & 59.69 & 66.14 & 28.73 & 30.26 & 34.87 \\
    &  \quad $+$ Shared Cache & 30.39 & 25.59 & 19.49 & 54.85 & 40.90 & 44.92 & 44.88 & 44.63 & 36.72 & 51.00 & 62.00 & 51.00 & 57.43 & 47.16 & 35.19 & 26.78 & 30.17 & 28.97 \\
    \bottomrule
    \end{tabular}
}

\caption{The detailed results on six different long sequence tasks, where the start size is set to $4$ and $l_s=256, k=768$ for all methods. Results are separately presented by grouping text with different source lengths.}
    \label{tab:start_size_details}
\end{table*}

\section{More Related Work}
\label{app:related_work}
\subsection{Long Sequence Processing}
Long sequence language modeling have attracted more and more research interests in recent years~\citep{tiezzi2024resurgence},  as large language models continue to advance~\citep{li2024fundamental}.
Various long document processing tasks are proposed to evaluate the long sequence modeling of language models~\citep{zhao2021ror, luo2021cooperative,bai2023longbench}.
Longformer, leveraging sparse self-attention pattern, save the memory cost to make the model process long document~\citep{beltagy2020longformer}.
Memorizing transformer uses a external memory to save the information during the long sequence modeling process~\citep{wu2022memorizing}.
Mistral applied Pre-fill and chunking sliding window methods to model longer sequences ~\citep{jiang2023mistral}.
State space models and their variations are also popular recently~\citep{gu2021efficiently, gu2023mamba,wang2022pretraining}.
Unlimitedformer wraps pretrained encoder-decoder transformer, and offloads the cross-attention computation to a single k-nearest-neighbor index, while the returned $k$NN distances are the attention dot-product scores~\citep{bertsch2024unlimiformer}.
~\citet{nawrot2024dynamic} propose to compress the key-value cache to make the model process longer sequences.
\citet{xiong2023effective} conduct continual pretraining from Llama 2~\citep{touvron2023llama} with longer training sequences and on a dataset where long texts
are upsampled.
Rotary Position Embedding and the positional interpolation based on it are also used enable the model process longer sequences~\citep{SU2024127063, chen2023extending}.
Text summarization has also been known by its relation with long sequence processing area~\citep{du2023domain, gao2024latent, li2024word}.
ReadAgent are proposed by using a large language model agent to process long sequences~\citep{lee2024humaninspired}.
\textsc{LongHeads} enhances the long-context processing of large language models by allowing multi-head attention to attend to selected important context chunks within the trained length~\citep{lu2024longheads}.
Infini-Transformer leverage a compressive memory between different context segment to achieve modeling long range text~\citep{munkhdalai2024leave}.
\citet{hwang2024transformerfam} propose TransformerFAM, a novel architecture with a feedback loop for attending to latent representations, enables Transformers to process indefinitely long sequences without additional weights.
\citet{zhang2024found} leverage plug-and-play positional encoding to make the model better collect the information in the middle of the document.

Except \textsc{LongHeads} which requires storing all the past key-value states, all the above needs further training to make the model able to handle the long sequence processing task. Our work do not need any training and can be applied directly to any open-source transformer-based large language models.

Retrieval-augmented generation techniques also share similar aspects with our methods. RAG techniques usually involve two steps: first, retrieving relevant information (usually a document) from a large database, and second, concatenating the document and the user query to enhance the performance of generating the response text. 
The similarity between our method and RAG methods mainly lies in the fact that our method can be applied to long document question-answering tasks, which is the typical form of the final step of the RAG methods. In this sense, our method is orthogonal to them, as it aims to improve the LLMs themselves and can handle documents that exceed the length limitations of LLMs in the RAG process. Hence, it is not appropriate to directly compare our methods to RAG techniques.

\subsection{State Eviction for Large Language Models}

~\citet{liu2024scissorhands} explore the persistence of importance hypothesis for the key-value cache of large language models. They establish that the key-value cache that useful for large language modeling are consistent for all the following text. Based on this, various methods that evicts the key-value cache during the language modeling has been proposed for improving the efficiency of the LLM inference.
~\citet{xiao2023efficient} propose that ``attention sink'' exists during the sequence processing of large language models. By keeping the key-value states of the initial tokens, and evict the key-value states out of a sliding window maintained for recent tokens, the model could maintain the perplexity while processing 1 million tokens.
~\citet{zhang2024h2o} use accumulative attention scores to evict the unnecessary key-value cache states.
\citet{oren2024transformers} uses the attention of the last token as a metric to evict the hidden states.
~\citet{ge2023model} profile all the attention heads and maintain different hidden states for different heads.
Attendre~\citep{yang2024attendre} brings the preference of future tokens into the state eviction process.

Besides inference-only state-eviction, a lot of methods also explore to learn to prune tokens during the training process in computer vision~\citep{wang2023zerotprune, kim2022learned, ye2021tr} or natural language processing~\citep{zhuang-wang-2019-token, frantar2023sparsegpt, yun-etal-2023-focus, anagnostidis2024dynamic}. There is also work that delete tokens from the discrete prompt~\citep{weston2023system}.

Compared to this paper, the previous work rarely focuses the state eviction technique on the long sequence modeling scenario and does not related to the specific optimization for the down-stream tasks.

\end{document}